\documentclass{article}
\usepackage[preprint]{neurips_2026}
\usepackage{graphicx}

\usepackage[utf8]{inputenc} % allow utf-8 input
\usepackage[T1]{fontenc}    % use 8-bit T1 fonts
\usepackage{hyperref}       % hyperlinks
\usepackage{url}            % simple URL typesetting
\usepackage{booktabs}       % professional-quality tables
\usepackage{amsfonts}       % blackboard math symbols
\usepackage{nicefrac}       % compact symbols for 1/2, etc.
\usepackage{microtype}      % microtypography
\usepackage{xcolor}         % colors
\usepackage{amsmath}
\usepackage{multirow}
\usepackage{array}
\usepackage{caption}
\usepackage{enumitem}
\usepackage{titlesec}

\hypersetup{colorlinks, linkcolor={red!55!black}, citecolor={blue!55!black}, urlcolor={blue!55!black}}

\title{Scaling Laws for HyperNetwork-Based Knowledge Injection in Large Language Models}

\author{%
  Nischay Dhankhar \\
  Nace AI \\
  \texttt{nischay@nace.ai}
  \And
  Dos Baha \\
  Nace AI \\
  \texttt{dos@nace.ai}
  \And
  Abulhair Saparov \\
  Nace AI \\
  Purdue University \\
  \texttt{asaparov@purdue.edu} 
}

\begin{document}

\maketitle

\begin{abstract}
Injecting factual knowledge into large language models (LLMs) reliably and at
scale remains an open challenge.

Hypernetworks provide a promising solution to large-scale knowledge injection.
Although hypernetworks are typically applied for \emph{test-time adaptation}, we explore their use in \emph{train-time knowledge injection}, where, given a large corpus of facts, we train a hypernetwork to generate a \emph{fixed} LoRA adapter that, when inserted into the target model, enable the model to answer questions about those facts.

In this work, we investigate
whether hypernetworks can be used to perform train-time knowledge injection and how this
ability varies with scale. The scaling behavior of hypernetworks themselves remains largely unstudied.

Our design decouples the injection capacity of
the hypernetwork from the general capability of the target model, enabling, for the
first time, a rigorous and controlled study of \textbf{scaling laws for hypernetwork
architectures} in the context of knowledge injection. We characterize how loss, reasoning accuracy, and out-of-distribution (OOD) generalization
vary as functions of hypernetwork depth, width, and target network size.
To this end, we construct a large-scale dataset, called \texttt{MegaWikiQA}, containing tens of millions of multi-hop question-answer examples across 39 knowledge domains constructed from examples in \texttt{Wikidata5M}. 

Our results reveal: 
\emph{(i)} hypernetwork-based injection exhibits broadly predictive power law
scaling along all architecture axes; and \emph{(ii)} hypernetworks are capable of reliable OOD generalization to unseen entities and relations at increasing scales, suggesting that
hypernetwork adaptation provides a promising alternative to other train-time adaptation methods such as 
LoRA finetuning and full fine-tuning, exhibiting steeper scaling 
exponents in all OOD evaluations. 
Together, these results establish hypernetworks as a principled and scalable
substrate for train-time knowledge injection, and provide the first empirically grounded
\textbf{scaling laws} to guide hypernetwork design for factual reasoning in large
language models.\footnote{Code and data are publicly available at \texttt{https://huggingface.co/collections/nace-ai/hypernetwork-datasets}.}
\end{abstract}

%--------------------------------------------------------
%  INTRODUCTION
%--------------------------------------------------------
\section{Introduction}
\label{sec:intro}

The ability to inject and generalize factual knowledge in large
language models (LLMs) is fundamental to their deployment in knowledge-intensive
settings, where models must reliably condition their outputs on specific facts
from a large fixed text corpus, particularly involving knowledge that may not be well-represented in pretraining.
For example, in medical applications, API-based LLM solutions are inappropriate due to the sensitivity of patient data, and adaptation of a local LLM is preferable ~\citep{DBLP:journals/corr/abs-2501-13687}. In specific domains, such as law and finance, questions may depend on domain-specific knowledge that is missing from the LLM ~\citep{DBLP:journals/csur/LingZLDZWCLCZZPMPCWLCCW26}, such as new financial regulations ~\citep{DBLP:journals/corr/abs-2306-06031}, enterprise internal policies ~\citep{DBLP:journals/corr/abs-2510-11588}, etc.

A natural approach to address this challenge is fine-tuning, but full fine-tuning can be prohibitively expensive.
Parameter-efficient fine-tuning (PEFT) methods such as LoRA~\citep{DBLP:conf/iclr/HuSWALWWC22}  can help to reduce the cost of fine-tuning, but both fine-tuning and PEFT methods are prone to catastrophic forgetting~\citep{DBLP:conf/iclr/KothaSR24, DBLP:conf/emnlp/Li0FT24} and can struggle with out-of-distribution (OOD) generalization~\citep{DBLP:conf/naacl/HajipourYSF24}.

\begin{figure}[t]
    \centering
    \includegraphics[trim={4em 1.35em 10em 7.6em},clip,width=0.95\textwidth]{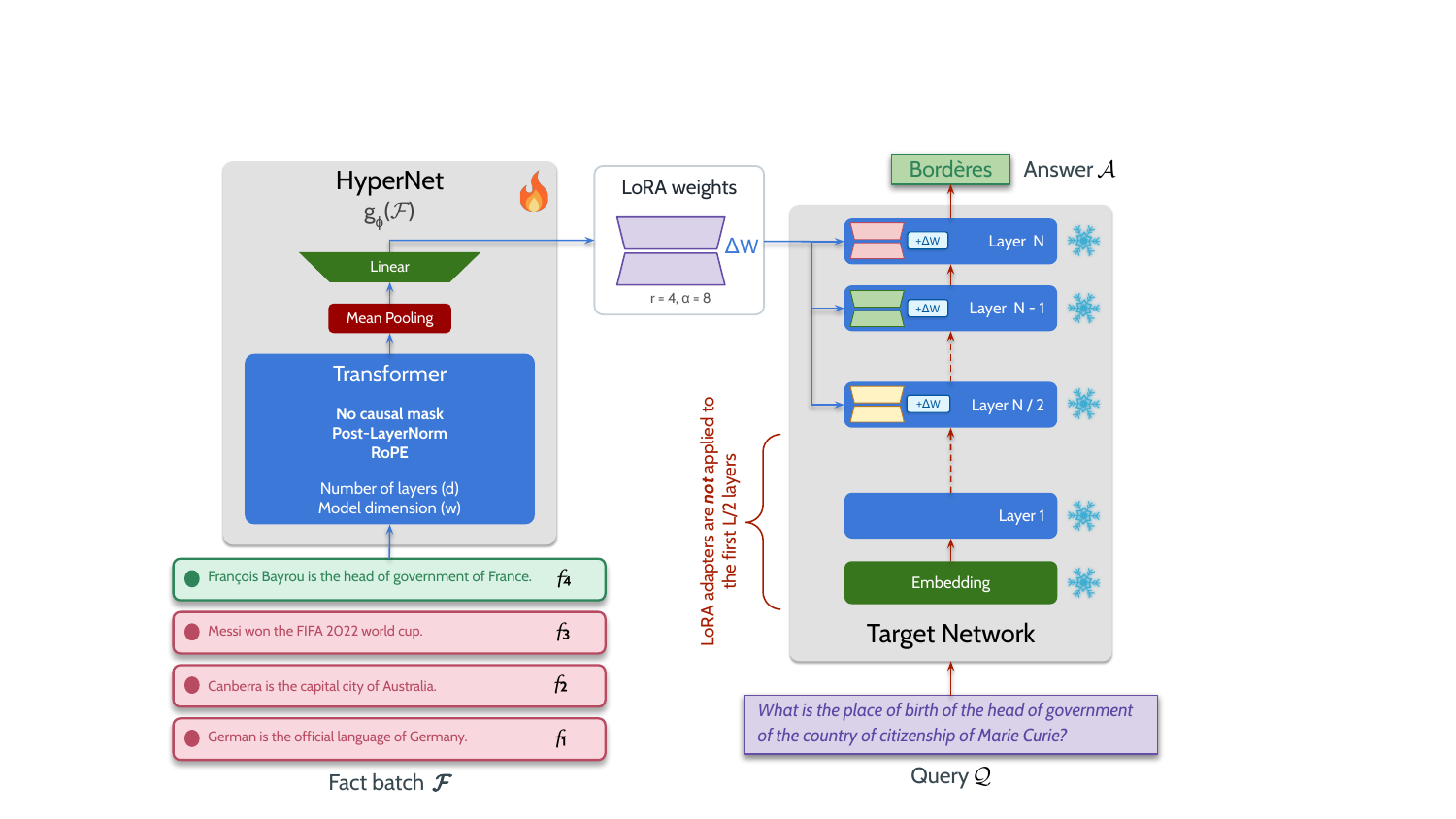}
    \vspace{-0.4em}
    \caption{Diagram of the proposed hypernetwork-based approach for knowledge injection, showcasing a single example. Only hypernetwork parameters are learned whereas target model parameters are frozen.}
    \label{fig:overview}
    \vspace{-0.4em}
\end{figure}

Due to these limitations, we explore an alternative and promising approach for
knowledge injection that does not share the same limitations. In this work, we study the injection of knowledge via hypernetworks~\citep{DBLP:conf/iclr/HaDL17} a paradigm in which a secondary network, conditioned on a set of injected facts, generates contextual weight adaptations (e.g., LoRA adapters) for the target model. This approach avoids modifying the base parameters of the target model, which can help to mitigate catastrophic forgetting and facilitate OOD generalization. Hypernetworks can incorporate additional information from their input which can lead to higher-quality representations and more compute-efficient learning of high-quality adapters for the target model. See Figure~\ref{fig:overview} in Section~\ref{sec:hypernet_architecture} for an overview.

We emphasize that our work focuses specifically on \emph{train-time knowledge 
internalization}: both the hypernetwork and all finetuning baselines we 
compare against are trained to compress a fixed fact corpus $\Omega$ into target 
model weights, and are evaluated on queries without any inference-time access 
to $\Omega$. This isolates the effect of the training-time adaptation 
mechanism, which is the object of study in our scaling laws.
Our novel approach is in stark contrast to the typical usage of hypernetworks for \emph{test-time adaptation}, which is more akin to \emph{knowledge 
editing}~\citep{DBLP:conf/iclr/MitchellLBFM22, DBLP:conf/nips/MengBAB22, 
DBLP:conf/iclr/MengSABB23, DBLP:journals/tacl/CohenBYGG24}, which we not consider in this paper. 

Despite their theoretical appeal, there exists no systematic characterization of the scaling behavior of hypernetwork-based knowledge injection. The central contribution of this paper is the first empirical and analytical
study of \textbf{scaling laws for hypernetwork-based knowledge injection}.
Concretely, we investigate how performance varies as a function of: (i) the
width of the hypernetwork; (ii) the depth of the hypernetwork; (iii) the size of the target
language model; and (iv) the quantity of injected
facts. In our experiments, we consider hypernetworks with sizes ranging from 167M to 2.8B parameters. In order to perform such scaling experiments, we require a large labeled dataset of question-answer pairs. To this end, we construct such a dataset, called \texttt{MegaWikiQA}, by extracting subject-relation-object triplets from \texttt{Wikidata5M}~\citep{DBLP:journals/tacl/WangGZZLLT21}, enabling controlled scaling over millions of facts with rigorous evaluation
of the model's multi-hop reasoning ability.

Our results reveal that hypernetwork-based knowledge injection exhibits smooth, predictable scaling behavior in contrast to the sharp failure modes documented for parametric editing. We further show that hypernetwork-generated adaptations generalize across unseen entities and relations, support multi-hop compositional queries, and
exhibit steeper OOD generalization scaling than both LoRA finetuning and 
full finetuning as the target model size increases, while achieving comparable 
performance on in-distribution validation.

We make the following core contributions in this work:
\begin{enumerate}[itemsep=-1pt,topsep=-7pt,leftmargin=1.8em]
  \item We provide the first systematic study of scaling laws for 
  hypernetwork-based knowledge injection along multiple axes: (1) 
  hypernetwork architecture parameters including transformer depth and 
  width; (2) the size of the target language model; and (3) the number 
  of injected facts (Section~\ref{sec:experiments}).

  \item We characterize the regimes where hypernetwork scaling yields 
  diminishing returns. We find that scaling the depth and width yields 
  comparable performance improvement, and that scaling the target model 
  offers substantially better improvement than scaling the hypernetwork 
  (Section~\ref{sec:target_scaling} and Table~\ref{tab:exponents}).

  \item We provide direct scaling law comparisons between hypernetwork 
  adaptation, LoRA finetuning, and full finetuning across target model 
  size. While finetuning methods scale slightly better on in-distribution 
  validation, the hypernetwork exhibits substantially steeper OOD 
  generalization scaling across all three OOD splits, and this advantage 
  grows with target model scale (Sections~\ref{sec:lora_target_scaling} 
  and \ref{sec:full_ft_target_scaling}).

  \item We release a reproducible large-scale benchmark, \texttt{MegaWikiQA}, 
  built on \texttt{Wikidata5M} with deterministic multi-hop question 
  generation and explicit OOD evaluation sets, suitable for knowledge 
  injection research, post-training, and evaluation of factual reasoning 
  in LLMs (Section~\ref{sec:dataset}).
\end{enumerate}

\section{Related Work}
\label{sec:related}

\paragraph{Fine-tuning and PEFT for Knowledge Injection.}
The term \emph{knowledge injection}\footnote{Following prior usage of the 
term~\citep{DBLP:conf/emnlp/OvadiaBME24, DBLP:conf/acl/WangTDWHJCJZ21, 
DBLP:conf/aaai/LiuZ0WJD020, DBLP:conf/acl/ZhangHLJSL19}, we use 
\emph{knowledge injection} to refer to the general task of teaching a 
pre-trained model a body of knowledge from a text corpus, regardless of 
whether that knowledge is entirely novel to the model. In particular, 
the model may already have partial exposure to the injected facts through 
pretraining; the goal is to reliably condition the model's outputs on the 
specified corpus.} was introduced in early work on injecting structured 
knowledge into pretrained language models, including 
K-BERT~\citep{DBLP:conf/aaai/LiuZ0WJD020}, 
K-Adapter~\citep{DBLP:conf/acl/WangTDWHJCJZ21}, and 
ERNIE~\citep{DBLP:conf/acl/ZhangHLJSL19}, and has since been widely adopted 
to describe fine-tuning-based approaches for incorporating knowledge from a corpus into models~\citep{DBLP:conf/emnlp/OvadiaBME24}.
Full fine-tuning is the most straightforward approach to knowledge 
injection, but is prohibitively expensive at scale and prone to 
catastrophic forgetting~\citep{DBLP:conf/iclr/KothaSR24, 
DBLP:conf/emnlp/Li0FT24}. Parameter-efficient fine-tuning (PEFT) methods 
such as LoRA~\citep{DBLP:conf/iclr/HuSWALWWC22} reduce adaptation 
cost but do not eliminate forgetting~\citep{DBLP:conf/emnlp/Li0FT24}, 
and struggle with out-of-distribution generalization to unseen entity 
and relation combinations~\citep{DBLP:conf/naacl/HajipourYSF24}. 
More fundamentally, injecting new factual knowledge through supervised 
fine-tuning is inherently difficult: new facts are learned significantly 
slower than facts that are consistent with pretraining, and as they are 
eventually learned, they increase the model's tendency to 
hallucinate~\citep{DBLP:conf/emnlp/GekhmanYAEFRH24}. 

\paragraph{Hypernetworks for Knowledge Injection.}
Hypernetworks~\citep{DBLP:conf/iclr/HaDL17} are networks whose outputs are the
weights of another network, enabling dynamic and input-conditioned parameterization.
Beyond knowledge editing, hypernetworks have been applied to parameter-efficient
adaptation across tasks and languages: notably, Hyper-X~\citep{DBLP:conf/emnlp/UstunBBNR22}
generates adapter module weights conditioned on joint task and language embeddings,
enabling zero-shot transfer to unseen task-language combinations without training
separate adapters per configuration. Two prior studies are most closely related to
ours in the context of knowledge editing. MEND~\citep{DBLP:conf/iclr/MitchellLBFM22}
trains a hypernetwork to transform the gradient of a language modeling loss on an
injected fact, applying the transformed gradient to update the base model.
Building on this, PropMEND~\citep{DBLP:journals/corr/abs-2506-08920} extends
MEND by modifying the meta-training objective to encourage multi-hop propagation
of injected knowledge, showing improved performance on non-verbatim propagation
questions. Critically, both MEND and PropMEND operate in gradient space: the
hypernetwork transforms a gradient signal, after which the resulting update is
applied to the base model parameters. These methods are prone to model instability when the number of injected facts grows beyond $\sim 1000$~\citep{DBLP:conf/acl/GuptaRA24}. Our approach departs from this
paradigm by using the hypernetwork to directly predict LoRA-style weight
adaptations~\citep{DBLP:conf/iclr/HuSWALWWC22} conditioned on the injected
facts at inference time, with no modification of base parameters, which helps to avoid instability.

\paragraph{Scaling Laws for Hypernetworks.}
The study of neural scaling laws gained prominence with~\citet{DBLP:journals/corr/abs-2001-08361},
who established power-law relationships between model size, dataset size, training compute,
and validation loss for autoregressive language models. Subsequent work extended scaling
analyses to specific capabilities including in-context learning \citep{DBLP:conf/nips/BrownMRSKDNSSAA20}, instruction
following \citep{DBLP:conf/nips/Ouyang0JAWMZASR22}, and reasoning \citet{DBLP:conf/nips/Wei0SBIXCLZ22}. More recently, \citet{DBLP:conf/icml/BethuneGBGCA25} extended scaling law analyses to the finetuning regime, deriving power-law relationships that quantify catastrophic forgetting as a function of model size, finetuning data volume, and the fraction of pretraining data injected into the finetuning mixture. The closest
prior work to ours is \citet{DBLP:conf/iclr/GuY25}, who show that hypernetwork performance
improves as the backbone foundation model is scaled up in the context of implicit
neural representation tasks. However, their scaling analysis is restricted to
varying the target network size, focuses on image and audio reconstruction rather
than knowledge injection, and does not explore scaling the hypernetwork
architecture itself. The scaling behavior of hypernetworks, which must
learn to generate weights for a separate frozen target network conditioned on
structured inputs, introduces qualitatively different questions: how does depth
versus width of the hypernetwork affect the quality of generated adaptations, and how does hypernetwork
capacity interact with target model size? To our knowledge, no prior work has
studied these questions empirically. Our work fills this gap by providing the
first empirical scaling laws for hypernetwork architectures in the context of
knowledge injection.

\paragraph{Knowledge Injection Datasets.}
Existing benchmarks for knowledge injection and editing vary significantly in scale
and construction methodology. \texttt{ZsRE}~\citep{DBLP:conf/conll/LevySCZ17} contains tens
of thousands of examples and predates modern LLMs, limiting its suitability for
large-scale evaluation. \texttt{UniEdit}~\citep{DBLP:journals/corr/abs-2505-12345} and
\texttt{WikiBigEdit}~\citep{DBLP:conf/icml/ThedeRBAH25} represent more recent efforts at
311K and 500K examples respectively, with broader domain coverage. However, both
are designed primarily for parametric inference-time knowledge editing rather than knowledge
injection. In inference-time knowledge editing, the task is to answer questions where each question is conditioned on new (possibly counterfactual) facts. Furthermore, past benchmarks rely on LLM-generated questions which may introduce noise and stylistic
bias. More broadly, none of these datasets are designed to support scaling experiments
across multiple orders of magnitude, where larger models may become increasingly sensitive to confounding effects such as noise and bias. We construct \texttt{MegaWikiQA} to fill this
gap, with fully deterministic grammar-based question generation, over 10 million
examples per number of hops, and explicit OOD domain splits for controlled evaluation.

\section{Dataset Construction: \texttt{MegaWikiQA}}
\label{sec:dataset}

A rigorous study of scaling laws for knowledge injection requires a dataset that is
large-scale, structurally rich, deterministic in its ground-truth labels, and capable
of supporting both single hop and multi-hop compositional evaluation. Support for multi-hop evaluation is critical because, in the knowledge injection setting, the model may be required to perform \emph{multi-hop and compositional reasoning over multiple injected facts} in order to answer a question, as opposed to recall individual memorized facts. Ideally, the dataset should contain several orders of magnitude in its number of examples to enable rigorous scaling experiments.
To that end, we construct our dataset, \texttt{MegaWikiQA}, entirely from \texttt{Wikidata5M}~\citep{DBLP:journals/tacl/WangGZZLLT21}, a large scale
knowledge graph containing approximately 4.6 million entities, 822 relations, and
over 22 million triplets of the form $(s, r, o)$, where $s$ is a subject entity,
$r$ is a relation, and $o$ is the object entity (e.g., $(\texttt{germany},\texttt{capital},\texttt{berlin})$ is a triplet representing the fact ``The capital of Germany is Berlin''). \texttt{Wikidata5M} satisfies all of the
above requirements: its scale provides ample coverage for training and evaluation
across 39 knowledge domains, its graph structure supports deterministic construction of multi-hop
reasoning examples on the scale of tens of millions, and its triplet format yields unambiguous ground-truth answers
for all question types we consider.

\subsection{Question Answer Generation}

Questions are generated by first performing a random walk in the \texttt{Wikidata5M} knowledge graph. We begin the walk with a seed triplet $(s_1, r_1, o_1)$, where $s_1$ is the subject entity, $o_1$ is the object entity, and $r_1$ is the relation. We then consider all neighboring edges---i.e., where the object of the first fact $o_1$ is the subject of the next fact---and select an edge uniformly at random. We repeat this process until we have a $k$-hop walk $(s_1, r_1, o_1, r_2, o_2, \ldots, r_k, o_k)$. We generate questions for multiple hop counts $k \in \{1, 2, 3, 4\}$. This process is repeated to generate multiple $k$-hop walks for each seed triplet, which we repeat again by considering every triplet in \texttt{Wikidata5M} as the seed triplet.

The $k$-hop walk is converted into a natural language question-answer pair using a \textbf{grammar-based approach}. Let $f$ be a (recursive) function that converts a given $k$-hop walk into a noun phrase:
\vspace{-0.3em}
\begin{align*}
    f(s_1&,r_1,o_1,r_2,o_2,\hdots,r_k,o_k) := \\[-0.3em]
    &\begin{cases}
        \text{``the country of citizenship of''} f(s_1,r_1,\hdots,r_{k-1},o_{k-1}) & \text{if } r_k = \texttt{citizen\_of\_country}, \\[-0.1em]
        \text{``the head of government of''} f(s_1,r_1,\hdots,r_{k-1},o_{k-1}) & \text{if } r_k = \texttt{head\_of\_govt}, \\[-0.1em]
        \text{``the place of birth of''} f(s_1,r_1,\hdots,r_{k-1},o_{k-1}) & \text{if } r_k = \texttt{place\_of\_birth}, \\[-0.4em]
        \hspace{10em}\vdots & \hspace{5em}\vdots
    \end{cases} \\[-1.6em]
\end{align*}
In the base case, where $k = 0$, the input sequence consists of a single entity $s_1$, which we verbalize: $f(s_1) = \text{``Marie Curie''}$ if $s_1 = \texttt{marie\_curie}$, etc.

We apply this function to convert the full $k$-hop walk into a noun phrase (e.g., ``the country of citizenship of Marie Curie''). Next, we convert it into a question (``What is the country of citizenship of Marie Curie?''). To perform this last conversion step, we manually curated and evaluated over 400 candidate question templates using a representative subset of triplets sampled from the dataset and constructed a deterministic
mapping from each of the 822 \texttt{Wikidata} relations to its most appropriate template.
Question templates have forms such as \textit{``What is $f(s_1,r_1,o_1,\hdots,r_k,o_k)$?''} and \textit{``Which is $f(s_1,r_1,o_1,\hdots,r_k,o_k)$?''}, with
the root selected based on the semantic type of the last object entity (e.g., \textit{``Who
is''} for objects with person type, \textit{``What is''} otherwise). The answer to the question is simply given by verbalizing the last object entity $o_k$ (e.g., $f(o_k) = \text{``France''}$ if $o_k = \texttt{france}$, etc).

To ensure the answers to our questions are unambiguious, we restrict our dataset to relations that are one-to-one or many-to-one, excluding one-to-many and many-to-many
relations whose answers are inherently non-deterministic. Although scaling experiments can be
performed with ambiguous question-answer pairs, this restriction reduces label
uncertainty and enables more reliable and interpretable evaluation using metrics
such as accuracy. 

\paragraph{Illustrative example.} Suppose we perform a random walk and obtain the sequence \textit{Marie Curie $\to$ country of citizenship $\to$ France
$\to$ head of government $\to$ Fran\c{c}ois Bayrou $\to$ place of birth $\to$
Bord\`eres}. Our procedure would then convert this into the question: \textit{``What is the place of birth of the
head of government of the country of citizenship of Marie Curie?''} with answer
\textit{Bord\`eres}. This approach is fully deterministic, requires no neural
generation at the question generation stage, and produces grammatically consistent
questions across all hop counts.

After performing this generation procedure over the full \texttt{Wikidata5M} dataset, we generated a multi-hop dataset containing approximately
\textbf{10 million examples per hop count} up to 4 hops. We classify each example into one of 39 domains using a two-stage classification pipeline described in Section~\ref{sec:domain}. We perform further filtering to remove examples on which the domain classification was uncertain, and to balance the number of examples across all hop counts. This resulted in a final dataset containing 1.25 million training samples stratified
across knowledge domains and reasoning complexity (i.e., number of hops).

\subsection{Fact Injection Protocol}

During training, our goal is to inject a large set of facts $\Omega$.
Each training example consists of an injected fact set $\mathcal{F} \subset \Omega$, a natural
language query $q$, and a ground truth answer $a$. The injected fact set always
contains one \emph{relevant fact}: from the $k$-hop sequence of facts that were used to generate the question-answer pair, the relevant fact is randomly selected from this sequence. The remaining $N - 1$ facts in $\mathcal{F}$ are \textbf{negative
facts} sampled uniformly at random from $\Omega$.

\subsection{Evaluation Protocol and OOD Splits}

We construct three disjoint evaluation sets to support comprehensive assessment
of knowledge injection and generalization.

\paragraph{In-distribution (ID) evaluation.} 10,000 randomly sampled examples
stratified by domain and number of hops, constructed to be disjoint from the training
set at the triplet level, i.e., no triplet appearing in the
evaluation set appears in the training set, and vice versa.

\paragraph{Out-of-distribution (OOD) evaluation.} 10,000 examples randomly drawn exclusively
from three held-out domains: \textit{philosophy}, \textit{linguistics}, and
\textit{civil engineering} as seen in Figure~\ref{fig:domain_accuracy}. These domains were selected due to their internal diversity and the fact that they were the most distinct as compared to the other domains. OOD
examples further include \textbf{rephrased questions} generated by prompting GPT 4.1~\citep{DBLP:journals/corr/abs-2303-08774} to paraphrase the example, with the aim to reduce potential overfitting to the highly-regular language produced by our grammar-based generation procedure, providing a stricter test
of generalization by controlling for confounding due to transfer learning of surface-level linguistic features. Finally, we include a \textbf{multiple-choice question (MCQ)} evaluation split, constructed from the same OOD examples where each question includes four answer choices: the correct entity and three distractor entities sampled uniformly at random from $\Omega$. This format tests whether the hypernetwork-adapted model can identify the correct answer under a fixed candidate set, providing a complementary evaluation signal that is robust to variations in generation style. Full dataset statistics, including split sizes and hop distributions, 
are summarized in Table~\ref{tab:dataset} in the Appendix.

\section{Method}
\label{sec:method}

\subsection{Problem Formulation}

We study the problem of knowledge injection of a large corpus of facts $\Omega$ into a frozen language model
$\mathcal{M}_\theta$ with fixed parameters $\theta$. Each training example
consists of a natural language query $q$ and a set of $N$ injected facts
$\mathcal{F} = \{f_1, \dots, f_N\} \subset \Omega$, where each fact $f_i$ is a natural
language verbalization of a triplet $(s, r, o) \in \Omega$. Exactly one
fact in $\mathcal{F}$ is \emph{relevant} to answering $q$, while the remaining
$N - 1$ facts are \emph{negatives} sampled uniformly at random from $\Omega$, encouraging the hypernetwork to
identify and utilize the relevant fact and improving generalization in the
presence of irrelevant information. The objective is to produce an answer $a$ consistent with the relevant fact
in $\mathcal{F}$, without modifying the base model parameters $\theta$ at
any point. Only the hypernetwork parameters $\phi$ are updated during
training; the target model $\mathcal{M}_\theta$ remains fully frozen
throughout both training and inference, which helps to mitigate the high cost of fine-tuning and may help to facilitate OOD generalization.

This formulation captures the core challenge of knowledge injection: the
hypernetwork must learn to identify and encode the relevant fact from a noisy
input context, and translate it into weight adaptations that steer the frozen
target model toward the correct answer. Unless otherwise stated, we fix $N = 4$
facts per example across all experiments, and study the effect of varying $N$
in dedicated experiments where we scale the number of facts (Section~\ref{sec:num_facts_scaling}).

\subsection{Hypernetwork Architecture}
\label{sec:hypernet_architecture}
We introduce a transformer-based hypernetwork $g_\phi$ with learnable parameters
$\phi$, initialized entirely from random weights (see Figure~\ref{fig:overview} 
for an architectural overview). A key design principle in our
study is that the hypernetwork is never initialized from pretrained weights,
as doing so would conflate the effect of pretraining with that of architectural
capacity, undermining the validity and controlled nature of our exploration on the effect of scale on hypernetwork performance. The hypernetwork
maps an input fact set $\mathcal{F}$ to a collection of LoRA-style weight
adaptations~\citep{DBLP:conf/iclr/HuSWALWWC22}, which are applied to the
frozen target model in the forward pass. We use LoRA rank $r = 4$ and scaling
factor $\alpha = 8$ across all experiments. Full details of the encoder
architecture, input encoding, LoRA weight generation, and target layer
selection are provided in Appendix~\ref{sec:hypernet_details}.

\paragraph{Architecture Scaling Axes.}
We systematically scale the transformer hypernetwork along three independent
axes to characterize the resulting scaling laws. For \textbf{depth scaling}
(Section~\ref{sec:depth_scaling}), we vary the number of transformer layers $L_{\text{HN}} \in
\{1, 2, 4, 8, 16\}$ while holding width fixed. For \textbf{width scaling}
(Section~\ref{sec:width_scaling}), we vary the hidden dimension $d_{\text{model}} \in \{32, 64,
128, 256, 512, 1024\}$ while holding depth fixed. For \textbf{fact count
scaling} (Section~\ref{sec:num_facts_scaling}), we vary the number of injected facts $N$ per example
while holding the hypernetwork architecture fixed, studying how the quantity
of available evidence per example affects injection performance. In all cases, the
hypernetwork is initialized from random weights. In each experiment, where we vary one of the above scaling variables, all other variables are held constant.

\section{Experiments}
\label{sec:experiments}

All scaling law experiments use \texttt{MegaWikiQA} as the knowledge source,
with training and evaluation splits constructed as in
Section~\ref{sec:dataset}. Unless otherwise stated, hypernetwork width scaling,
depth scaling, and fact count scaling experiments use
\texttt{Qwen2.5-1.5B-Instruct} as the frozen target model. %Target model scaling holds
%the hypernetwork architecture fixed and varies the target model across the
%\texttt{Qwen2.5} family.
We choose the \texttt{Qwen2.5} family since it offers a wide range of
model sizes within a consistent architecture, enabling controlled scaling
comparisons; newer families such as \texttt{Qwen3} have fewer variants and less
consistent training across models, making them less suitable for scaling analysis.

We evaluate all models on four metrics: in-distribution validation loss (ID validation),
OOD loss on held-out domains (OOD non-rephrased), OOD loss on rephrased questions on held-out domains (OOD rephrased), and OOD loss on multiple-choice questions on held-out domains (OOD MCQ), where each OOD split tests a qualitatively distinct form
of generalization as described in Section~\ref{sec:dataset}. Power-law fits of
the form $\mathcal{L} = a x^b$ are estimated via least-squares regression
in log-log space using the final epoch loss, where $x$ is the scaling variable of interest (e.g., hypernetwork width, etc).

%We note that, unlike scaling laws for autoregressive language modeling, there is no notable point during training at which the loss function stops decreasing quickly. Instead, the loss function seems to more gradually approach the minimum over the course of training (Appendix~\ref{sec:training_trajectories}).
%For each experiment, we perform training for sufficiently many epochs until the loss function stops decreasing quickly, and we continue training for a few additional epochs to observe the subsequent training dynamics.
%As such, for each experiment, we perform training for 5 epochs, which we observed to be sufficient for convergence across experiments.\footnote{Except in the depth scaling experiment, where we perform training for 6 epochs.}

\subsection{Hypernetwork Width Scaling}
\label{sec:width_scaling}

\begin{figure}
    \centering
    \includegraphics[width=\linewidth]{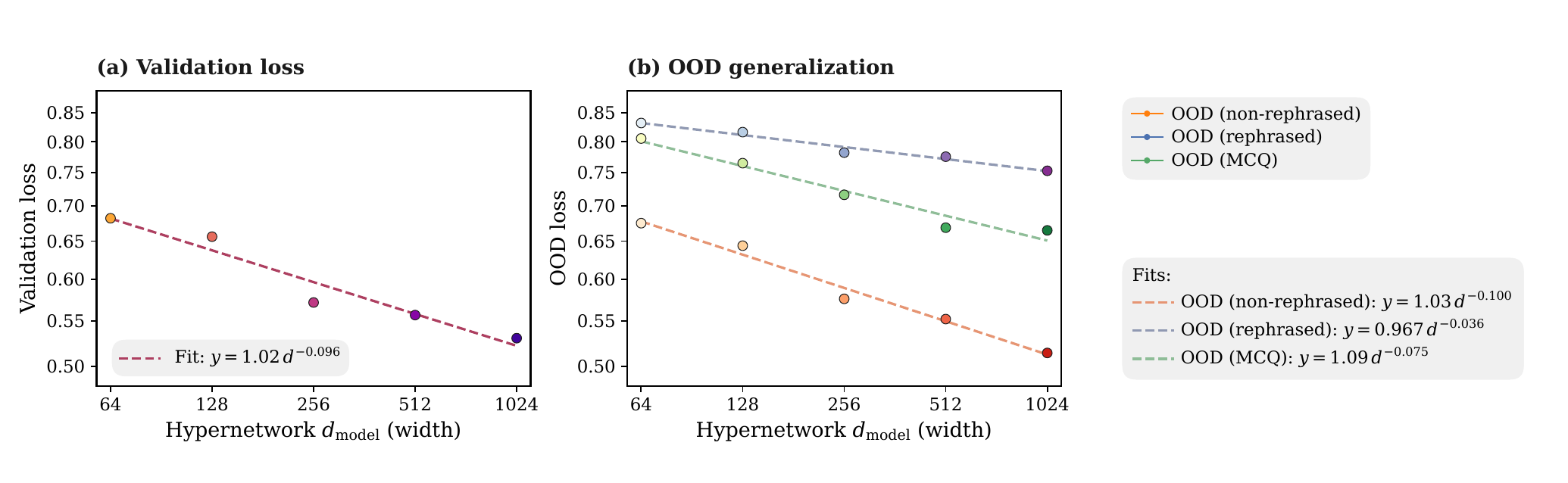}
\vspace{-1.8em}
    \caption{Hypernetwork width scaling results, showing final epoch loss vs.\ 
    $d_{\text{model}}$ on a log-log scale. \textbf{(a)} Validation loss follows 
    a power-law fit $\mathcal{L}_{\text{val}} = 1.02 \cdot d^{-0.096}$. 
    \textbf{(b)} All three OOD metrics improve with width, with fitted exponents 
    $-0.100$ (OOD non-rephrased), $-0.036$ (OOD rephrased), and $-0.075$ (OOD MCQ).}
    \label{fig:width_loglog}
\vspace{-1.1em}
\end{figure}

We vary the hypernetwork hidden dimension $d_{\text{model}} \in \{64, 128, 256,
512, 1024\}$, holding the number of transformer layers fixed and all other
hyperparameters constant. Figure~\ref{fig:width_loglog} shows the final epoch
loss vs $d_{\text{model}}$ on a log-log scale. Loss trajectories during training are provided in Appendix~\ref{sec:training_trajectories} 
(Figure~\ref{fig:width_traj}).

All four evaluation metrics follow smooth power-law relationships vs hypernetwork width.
The validation loss exponent ($-0.096$) is comparable to the OOD non-rephrased
exponent ($-0.100$), indicating that width scaling benefits both in-distribution
and OOD recall roughly equally, if the questions have similar phrasing. However, the OOD rephrased exponent
($-0.036$) is substantially flatter, suggesting that width alone does not
substantially improve robustness to linguistic surface variation. The OOD MCQ
exponent ($-0.075$) falls between the two, reflecting that multiple-choice
reformulation is a moderately harder generalization target as compared to open-ended generation.

\subsection{Hypernetwork Depth Scaling}
\label{sec:depth_scaling}

\begin{figure}
    \centering
    \includegraphics[width=\linewidth]{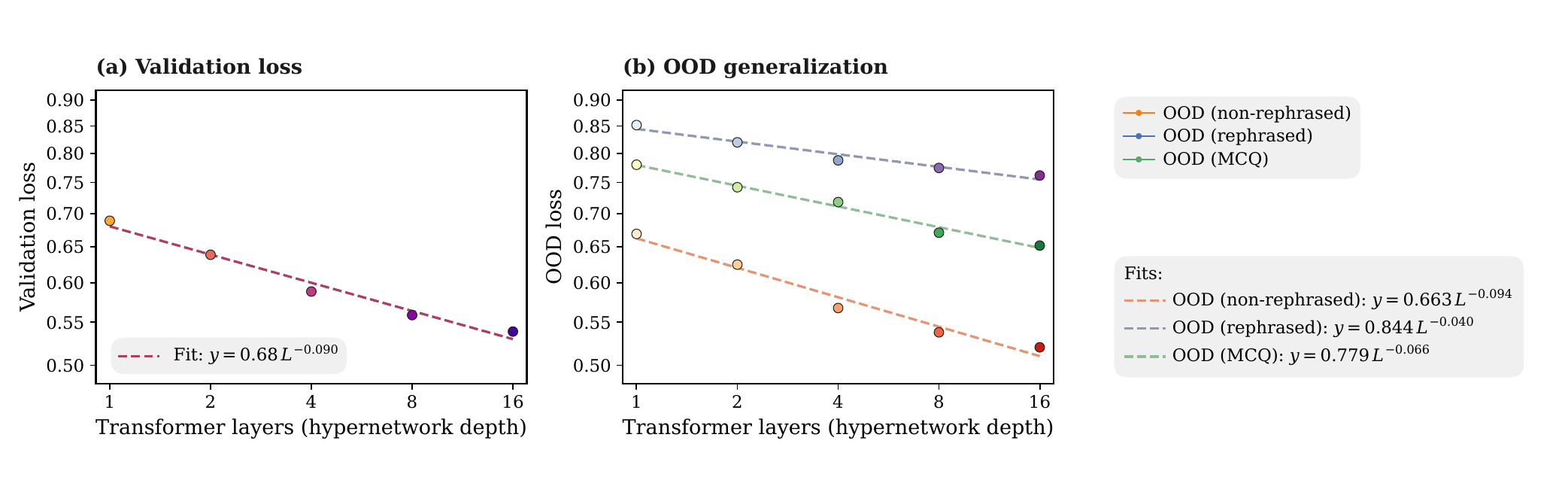}
\vspace{-1.7em}
    \caption{Hypernetwork depth scaling results, showing final epoch loss vs.\ 
    number of transformer layers $L_{\text{HN}}$ on a log-log scale. 
    \textbf{(a)} Validation loss follows a power-law fit 
    $\mathcal{L}_{\text{val}} = 0.677 \cdot L^{-0.088}$. \textbf{(b)} All three 
    OOD metrics improve with depth, with fitted exponents $-0.096$ (OOD 
    non-rephrased), $-0.042$ (OOD rephrased), and $-0.063$ (OOD MCQ).}
    \label{fig:depth_loglog}
\vspace{-0.5em}
\end{figure}

We scale the number of transformer layers in the hypernetwork across
$L_{\text{HN}} \in \{1, 2, 4, 8, 16\}$, holding $d_{\text{model}} = 512$ and
all other hyperparameters fixed. Figure~\ref{fig:depth_loglog} shows the final
epoch loss against $L_{\text{HN}}$ on a log-log scale for all four evaluation
metrics. Loss trajectories during training are shown in Figure~\ref{fig:depth_traj} in Appendix~\ref{sec:training_trajectories}. 

All four evaluation metrics follow smooth power-law relationships with
hypernetwork depth. The fitted exponents are $-0.088$ for validation loss,
$-0.096$ for OOD non-rephrased, $-0.042$ for OOD rephrased, and $-0.063$ for OOD
MCQ. As in the case with width scaling, the exponent on in-distribution and OOD non-rephrased metrics is steeper as compared to OOD rephrased, suggesting that increasing hypernetwork depth does not exhibit better 
surface-level linguistic generalization as compared to increasing width.
Comparing the width exponents to the depth exponents reported in
Section~\ref{sec:depth_scaling}, we observe broadly similar scaling rates
($-0.096$ vs $-0.088$ for validation loss), indicating that hypernetwork depth and width
contribute comparably to overall knowledge injection performance within the ranges studied.

\subsection{Target Model Scaling}
\label{sec:target_scaling}

\begin{figure}[t]
    \centering
    \includegraphics[width=\linewidth]{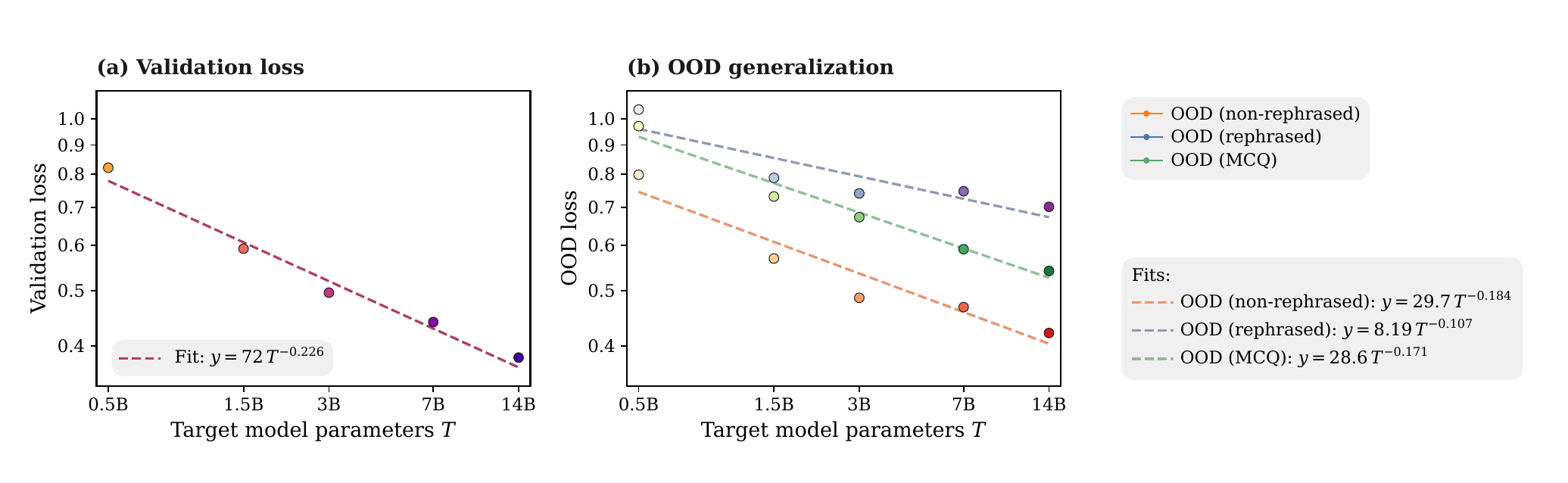}
\vspace{-1.7em}
    \caption{Target model scaling results, showing final epoch loss vs.\ target 
model parameter count $T$ on a log-log scale. \textbf{(a)} Validation loss 
follows a power-law fit $\mathcal{L}_{\text{val}} = 72 \cdot T^{-0.226}$. 
\textbf{(b)} All three OOD metrics improve with target model size, with 
fitted exponents $-0.184$ (OOD non-rephrased), $-0.107$ (OOD rephrased), 
and $-0.171$ (OOD MCQ).}
    \label{fig:target_loglog}
\vspace{-1.0em}
\end{figure}

Figure~\ref{fig:target_loglog} shows the final epoch loss against
target model size on a log-log scale. Loss trajectories are provided in Appendix~\ref{sec:training_trajectories} 
(Figure~\ref{fig:target_traj}).

Target model scaling yields the steepest power-law exponents across all axes
studied: $-0.226$ for validation loss and $-0.184$ for OOD non-rephrased loss,
compared to $-0.088$ and $-0.096$ for depth and width scaling respectively. This
indicates that scaling the target model yields substantially larger absolute
gains per unit compute as compared to scaling the hypernetwork architecture alone,
and suggests a practical guideline: when a fixed compute budget must be
allocated between hypernetwork capacity and target model size, the latter
offers a more favorable return.

The 0.5B target model lies noticeably above the power-law fit on the right-side of the figure, suggesting the existence of a minimum target model capacity threshold
below which hypernetwork-generated LoRA adaptations cannot be effectively
leveraged.
%In contrast, the 3B and 7B models converge to nearly identical final
%OOD loss, indicating a plateau in this size range that the 14B model breaks
%through.
%Across all three OOD evaluation splits, the ranking by target model
%size is preserved throughout training with no crossover, confirming that the
%scaling trend is robust to the form of OOD evaluation.

\subsection{Fact Count Scaling}
\label{sec:num_facts_scaling}

\begin{figure}[t]
    \centering
    \includegraphics[width=\linewidth]{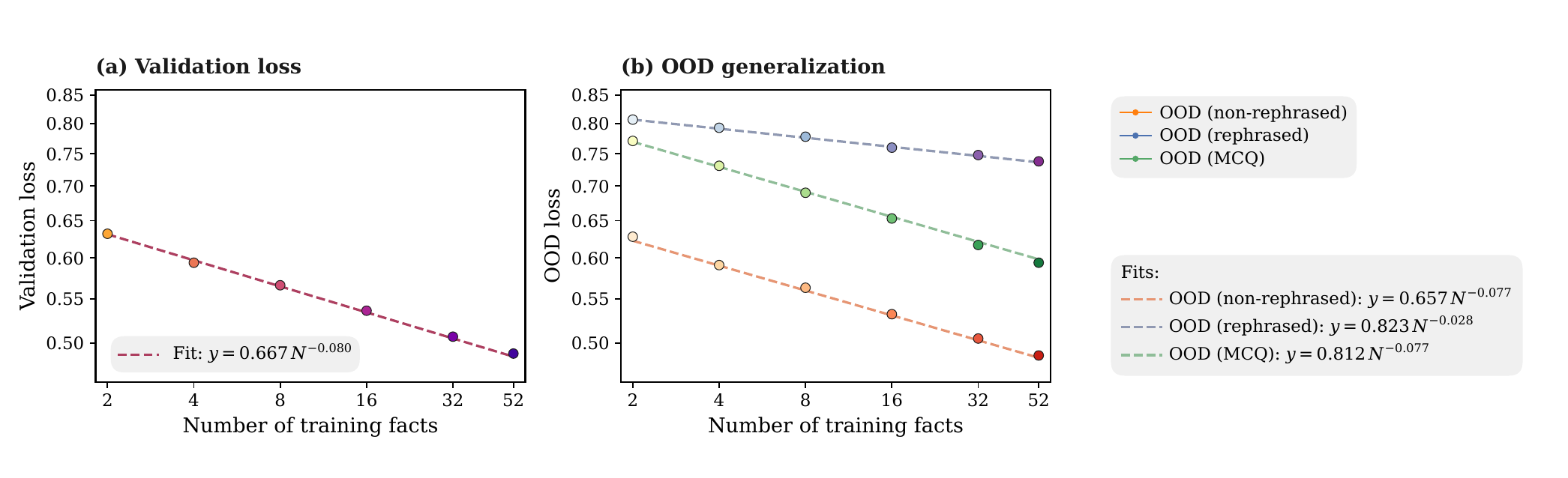}
\vspace{-1.8em}
    \caption{Fact count scaling results, showing final epoch loss vs.\ num. 
    of injected facts $N$ per example on a log-log scale. \textbf{(a)} 
    Validation loss follows a power-law 
    $\mathcal{L}_{\text{val}} = 0.667 \cdot N^{-0.080}$. \textbf{(b)} OOD 
    metrics improve with fact count. Fitted exponents: $-0.077$ (OOD 
    non-rephrased), $-0.028$ (OOD rephrased), $-0.077$ (OOD MCQ).}
    \label{fig:facts_loglog}
\vspace{-1.4em}
\end{figure}

We vary the number of training facts $N \in \{2, 4, 8, 16, 32, 52\}$ injected
per example, holding the hypernetwork architecture and target model fixed.\footnote{The maximum number of facts that we consider is 52 since the hypernetwork context is limited to 512 tokens.}
Figure~\ref{fig:facts_loglog} shows the final epoch loss against $N$ on a
log-log scale. Loss trajectories during training are provided in Appendix~\ref{sec:training_trajectories} 
(Figure~\ref{fig:facts_traj}).

Performance improves as the number of injected facts per example
grows, following power-law trends across all metrics. The validation loss
exponent ($-0.080$) and OOD non-rephrased exponent ($-0.077$) are modest but stable,
indicating that providing more factual context reliably aids the
hypernetwork's ability to identify and encode the relevant fact. The OOD
rephrased exponent ($-0.028$) is again the flattest, consistent with the
pattern observed in depth and width scaling. %, and further supporting the conclusion that OOD robustness to linguistic surface variation is a distinct and harder generalization challenge.
The OOD MCQ exponent ($-0.077$) matches
the OOD non-rephrased exponent closely, suggesting that multiple-choice and
free-form evaluation are comparably affected by the quantity of available
evidence.

\subsection{LoRA Scaling}
\label{sec:lora_target_scaling}

\begin{figure}[h]
    \centering
    \includegraphics[width=\linewidth]{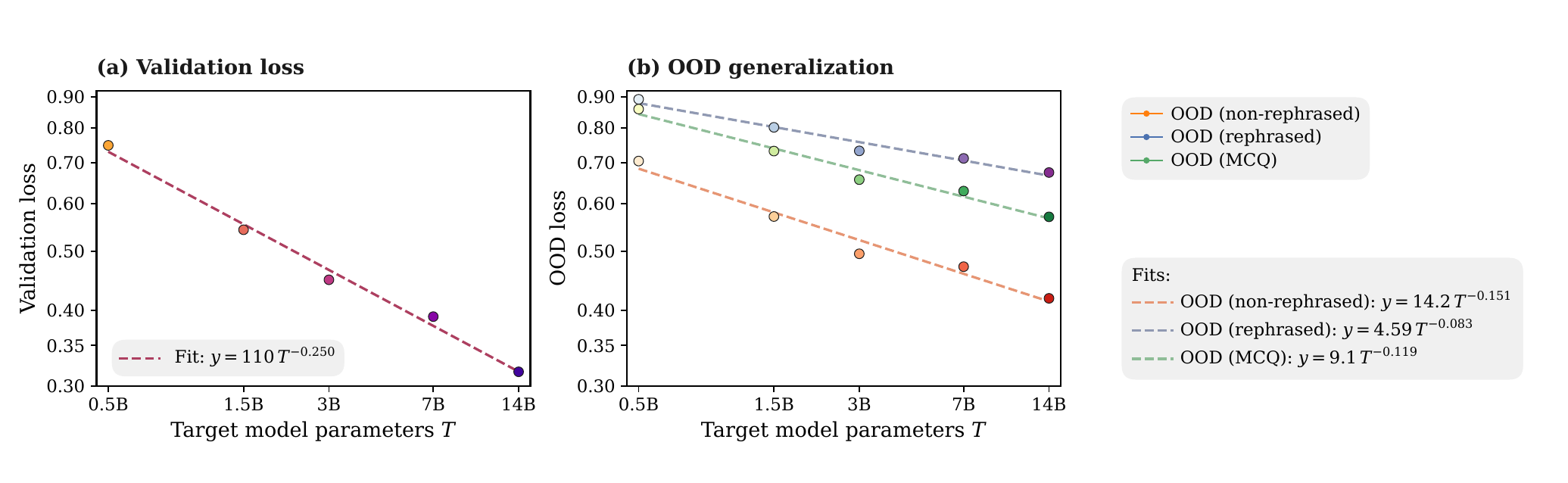}
    \vspace{-1.8em}
    \caption{LoRA target model scaling results ($r = 16, \alpha = 32$), 
    showing final epoch loss vs.\ target model parameter count $T$ on a 
    log-log scale. \textbf{(a)} Validation loss follows a power-law fit 
    $\mathcal{L}_{\text{val}} = 110 \cdot T^{-0.250}$. \textbf{(b)} OOD 
    metrics improve with target model size, with fitted exponents 
    $-0.151$ (OOD non-rephrased), $-0.083$ (OOD rephrased), and $-0.119$ 
    (OOD MCQ).}
    \label{fig:lora_target_loglog}
    \vspace{-1.0em}
\end{figure}

To complete the head-to-head comparison, we also experiment with scaling the target model 
under LoRA finetuning with fixed adapter configuration $r = 16, \alpha = 32$, 
across the same five \texttt{Qwen2.5} sizes used in the hypernetwork target 
scaling experiments. Figure~\ref{fig:lora_target_loglog} shows the final 
epoch loss against target model size on a log-log scale.

Comparing LoRA target scaling to hypernetwork target scaling reveals a 
consistent pattern: LoRA achieves a marginally steeper ID validation loss 
exponent ($-0.250$ vs.\ $-0.226$ for the hypernetwork), but its OOD 
scaling exponents are substantially flatter across all three OOD splits 
($-0.151$ vs.\ $-0.184$ on non-rephrased, $-0.083$ vs.\ $-0.107$ on 
rephrased, and $-0.119$ vs.\ $-0.171$ on MCQ). In other words, increasing target 
model size leads to a larger improvement in LoRA fitting the training distribution as compared to the hypernetwork fitting the training distribution, but it leads to a substantially larger improvement in the hypernetwork's OOD generalization as compared to LoRA. This gap is 
consistent with the OOD advantage observed in the comparison at fixed 
target model size (Section~\ref{sec:comparison}) and demonstrates that 
the OOD advantage of hypernetwork-based injection persists and, in fact, 
widens as the target model is scaled.

\subsection{Full Finetuning Scaling}
\label{sec:full_ft_target_scaling}

\begin{figure}[h]
    \centering
    \includegraphics[width=\linewidth]{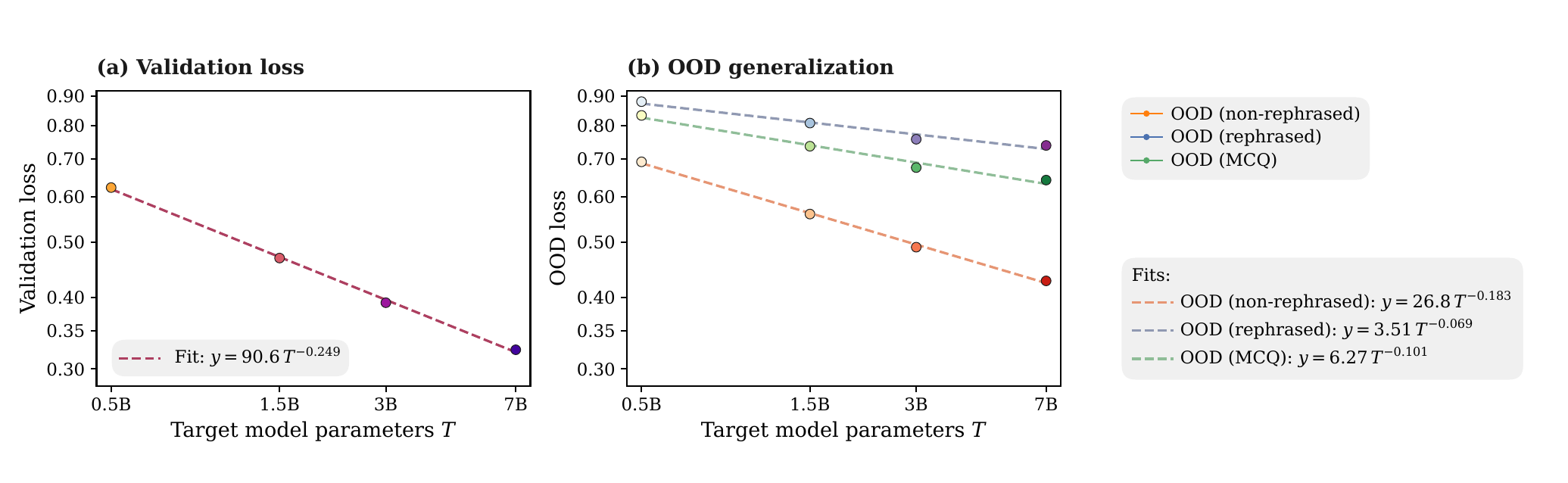}
    \vspace{-1.8em}
    \caption{Full finetuning target model scaling results, showing final 
    epoch loss vs.\ target model parameter count $T$ on a log-log scale. 
    \textbf{(a)} Validation loss follows a power-law fit 
    $\mathcal{L}_{\text{val}} = 90.6 \cdot T^{-0.249}$. \textbf{(b)} OOD 
    metrics improve with target model size, with fitted exponents 
    $-0.183$ (OOD non-rephrased), $-0.069$ (OOD rephrased), and $-0.101$ 
    (OOD MCQ). The 14B point was not evaluated due to compute constraints.}
    \label{fig:full_ft_target_loglog}
    \vspace{-1.0em}
\end{figure}

For completeness, we also experiment with scaling the target model under full finetuning, 
where all target model parameters are updated during training. Due to 
compute constraints, we cover four \texttt{Qwen2.5} sizes for full 
finetuning: 0.5B, 1.5B, 3B, and 7B\footnote{Full finetuning for the 14B configuration was infeasible 
under our budget.}. Figure~\ref{fig:full_ft_target_loglog} shows the final 
epoch loss against target model size on a log-log scale.

Full finetuning target scaling exhibits a pattern qualitatively similar 
to LoRA finetuning: a steep ID validation loss exponent ($-0.249$, nearly 
matching LoRA's $-0.250$) but flatter OOD loss exponents ($-0.183, -0.069, 
-0.101$) than the hypernetwork ($-0.184, -0.107, -0.171$). While full 
finetuning approximately matches the hypernetwork on OOD non-rephrased 
scaling, it lags substantially on OOD rephrased and OOD MCQ, the two 
metrics that test robustness to linguistic and format variation. If the 
observed power-law trends hold, the OOD advantage of the hypernetwork 
over full finetuning would be expected to widen further at 14B, 
consistent with the same pattern observed in the LoRA scaling 
comparison (Section~\ref{sec:lora_target_scaling}).

\subsection{Summary of Scaling Exponents}

Table~\ref{tab:exponents} summarizes the power-law exponents across all four
scaling axes and all four evaluation metrics. Target model scaling consistently
yields the steepest exponents, followed by depth and width scaling at comparable rates,
with fact count scaling yielding the most modest improvements. Across all axes,
the OOD rephrased metric exhibits the flattest scaling, suggesting that
robustness to linguistic variation is not easily addressed by scaling any single
architectural or data dimension.

\begin{table}[h]
\vspace{-1.2em}
    \caption{Summary of power-law scaling exponents across all target-scaling 
    axes and evaluation metrics. More negative values indicate steeper (more 
    favorable) scaling. HN denotes hypernetwork, FT denotes finetuning. LoRA 
    rank scaling is reported separately in Appendix~\ref{sec:lora_rank_scaling} 
    since it is better fit by a power-law with an additive constant.}
    \label{tab:exponents}
    \centering
    \small
    \begin{tabular}{lcccc}
        \toprule
        Scaling Axis & ID validation & OOD non-rephrased & OOD rephrased & OOD MCQ \\
        \midrule
        HN width ($d_{\text{model}}$)      & $-0.096$ & $-0.100$ & $-0.036$ & $-0.075$ \\
        HN depth ($L_{\text{HN}}$)         & $-0.088$ & $-0.096$ & $-0.042$ & $-0.063$ \\
        HN target model size ($T$)         & $-0.226$ & $\mathbf{-0.184}$ & $\mathbf{-0.107}$ & $\mathbf{-0.171}$ \\
        Fact count ($N$)                   & $-0.080$ & $-0.077$ & $-0.028$ & $-0.077$ \\
        \midrule
        LoRA FT target model size ($T$)    & $\mathbf{-0.250}$ & $-0.151$ & $-0.083$ & $-0.119$ \\
        Full FT target model size ($T$)    & $-0.249$ & $-0.183$ & $-0.069$ & $-0.101$ \\
        \bottomrule
    \end{tabular}
\vspace{-0.5em}
\end{table}

\subsection{Comparison of Scaling Laws Across Adaptation Methods}
\label{sec:comparison}

Our target model scaling experiments across the three train-time adaptation 
methods (hypernetwork, LoRA finetuning, and full finetuning) enable a direct 
comparison of their scaling behavior as target model capacity grows. 
Table~\ref{tab:exponents} reports the fitted power-law exponents for all 
three methods across all four evaluation metrics.

Two consistent patterns emerge. First, on in-distribution validation, 
finetuning methods scale slightly better than the hypernetwork: LoRA 
finetuning ($-0.250$) and full finetuning ($-0.249$) achieve nearly 
identical and steeper exponents than the hypernetwork ($-0.226$). This 
reflects the ability of direct parameter updates, whether full or 
low-rank, to fit the training distribution more aggressively as target 
capacity grows. Second, on all three OOD splits, the hypernetwork 
exhibits the steepest scaling among the three methods, most notably on 
OOD rephrased ($-0.107$ vs.\ $-0.083$ for LoRA and $-0.069$ for full FT) 
and OOD MCQ ($-0.171$ vs.\ $-0.119$ and $-0.101$). On OOD non-rephrased, 
the hypernetwork ($-0.184$) narrowly leads full FT ($-0.183$) and LoRA 
($-0.151$).

The gap between hypernetwork and finetuning OOD scaling grows monotonically 
with target model size, meaning that the OOD advantage of hypernetwork-based 
adaptation is not a fixed offset but widens as target models become larger. 
This is arguably the most important finding of our scaling law comparison: 
if the goal is train-time knowledge injection that generalizes robustly 
outside the training distribution, the hypernetwork paradigm becomes 
increasingly favorable at scale, precisely the regime where deployment 
matters most.

\section{Conclusion and Future Work}

We introduce \texttt{MegaWikiQA} and conduct the first systematic study of 
scaling laws for hypernetwork-based knowledge injection, characterizing 
power-law improvements across hypernetwork width, depth, target model size, 
and number of injected facts. Our results show that target model scaling 
yields the strongest gains, that hypernetworks generalize more effectively 
to OOD settings than other finetuning methods under matched capacity, and that 
these trends hold across unseen domains, rephrased queries, and 
query formats. 

A practical concern that emerges at larger scales is that the hypernetwork
itself can become prohibitively large relative to the target model it is
adapting. In our highest-capacity experiments, the hypernetwork reached
approximately 2.5B parameters while the target model was only 1.5B parameters,
making the hypernetwork nearly as large as the target model. This
parameter overhead significantly limits the practical deployability of large
hypernetworks and motivates future work on more parameter-efficient hypernetwork
architectures, such as shared weight generation across layers, low-rank
hypernetwork designs, or distillation of large hypernetworks into smaller ones
after training. Our target model scaling experiments cover the \texttt{Qwen2.5} family up to 14B
parameters. Extending this analysis to frontier-scale models (70B and beyond)
remains an important open question, as the relationship between hypernetwork
capacity and target model size may exhibit qualitatively different behavior at
much larger scales. Our current evaluation covers single-hop and shallow multi-hop queries from
\texttt{Wikidata5M}. A natural extension is to study whether hypernetwork-based injection
supports deep multi-hop compositional reasoning and long-horizon reasoning, where answering a query requires
the simultaneous use of many more injected facts. Whether inference-time weight generation
can scale to this regime, and how the required hypernetwork capacity grows with
reasoning depth, are open questions that we leave to future work.

\bibliography{sources}

@inproceedings{DBLP:conf/nips/MengBAB22,
  author       = {Kevin Meng and
                  David Bau and
                  Alex Andonian and
                  Yonatan Belinkov},
  editor       = {Sanmi Koyejo and
                  S. Mohamed and
                  A. Agarwal and
                  Danielle Belgrave and
                  K. Cho and
                  A. Oh},
  title        = {Locating and Editing Factual Associations in {GPT}},
  booktitle    = {Advances in Neural Information Processing Systems 35: Annual Conference
                  on Neural Information Processing Systems 2022, NeurIPS 2022, New Orleans,
                  LA, USA, November 28 - December 9, 2022},
  year         = {2022},
  url          = {http://papers.nips.cc/paper\_files/paper/2022/hash/6f1d43d5a82a37e89b0665b33bf3a182-Abstract-Conference.html},
  bibsource    = {dblp computer science bibliography, https://dblp.org}
}

@article{DBLP:journals/tacl/CohenBYGG24,
  author       = {Roi Cohen and
                  Eden Biran and
                  Ori Yoran and
                  Amir Globerson and
                  Mor Geva},
  title        = {Evaluating the Ripple Effects of Knowledge Editing in Language Models},
  journal      = {Trans. Assoc. Comput. Linguistics},
  volume       = {12},
  pages        = {283--298},
  year         = {2024},
  url          = {https://doi.org/10.1162/tacl\_a\_00644},
  doi          = {10.1162/TACL\_A\_00644},
  bibsource    = {dblp computer science bibliography, https://dblp.org}
}

@inproceedings{DBLP:conf/iclr/MengSABB23,
  author       = {Kevin Meng and
                  Arnab Sen Sharma and
                  Alex J. Andonian and
                  Yonatan Belinkov and
                  David Bau},
  title        = {Mass-Editing Memory in a Transformer},
  booktitle    = {The Eleventh International Conference on Learning Representations,
                  {ICLR} 2023, Kigali, Rwanda, May 1-5, 2023},
  publisher    = {OpenReview.net},
  year         = {2023},
  url          = {https://openreview.net/forum?id=MkbcAHIYgyS},
  bibsource    = {dblp computer science bibliography, https://dblp.org}
}

@inproceedings{DBLP:conf/acl/GuptaRA24,
  author       = {Akshat Gupta and
                  Anurag Rao and
                  Gopala Anumanchipalli},
  editor       = {Lun{-}Wei Ku and
                  Andre Martins and
                  Vivek Srikumar},
  title        = {Model Editing at Scale leads to Gradual and Catastrophic Forgetting},
  booktitle    = {Findings of the Association for Computational Linguistics, {ACL} 2024,
                  Bangkok, Thailand and virtual meeting, August 11-16, 2024},
  series       = {Findings of {ACL}},
  pages        = {15202--15232},
  publisher    = {Association for Computational Linguistics},
  year         = {2024},
  url          = {https://doi.org/10.18653/v1/2024.findings-acl.902},
  doi          = {10.18653/V1/2024.FINDINGS-ACL.902},
  bibsource    = {dblp computer science bibliography, https://dblp.org}
}

@inproceedings{DBLP:conf/iclr/HaDL17,
  author       = {David Ha and
                  Andrew M. Dai and
                  Quoc V. Le},
  title        = {HyperNetworks},
  booktitle    = {5th International Conference on Learning Representations, {ICLR} 2017,
                  Toulon, France, April 24-26, 2017, Conference Track Proceedings},
  publisher    = {OpenReview.net},
  year         = {2017},
  url          = {https://openreview.net/forum?id=rkpACe1lx},
  bibsource    = {dblp computer science bibliography, https://dblp.org}
}

@inproceedings{DBLP:conf/iclr/MitchellLBFM22,
  author       = {Eric Mitchell and
                  Charles Lin and
                  Antoine Bosselut and
                  Chelsea Finn and
                  Christopher D. Manning},
  title        = {Fast Model Editing at Scale},
  booktitle    = {The Tenth International Conference on Learning Representations, {ICLR}
                  2022, Virtual Event, April 25-29, 2022},
  publisher    = {OpenReview.net},
  year         = {2022},
  url          = {https://openreview.net/forum?id=0DcZxeWfOPt},
  bibsource    = {dblp computer science bibliography, https://dblp.org}
}

@article{DBLP:journals/corr/abs-2506-08920,
  author       = {Zeyu Leo Liu and
                  Greg Durrett and
                  Eunsol Choi},
  title        = {PropMEND: Hypernetworks for Knowledge Propagation in LLMs},
  journal      = {CoRR},
  volume       = {abs/2506.08920},
  year         = {2025},
  url          = {https://doi.org/10.48550/arXiv.2506.08920},
  doi          = {10.48550/ARXIV.2506.08920},
  eprinttype   = {arXiv},
  eprint       = {2506.08920},
  bibsource    = {dblp computer science bibliography, https://dblp.org}
}

@inproceedings{DBLP:conf/iclr/HuSWALWWC22,
  author       = {Edward J. Hu and
                  Yelong Shen and
                  Phillip Wallis and
                  Zeyuan Allen{-}Zhu and
                  Yuanzhi Li and
                  Shean Wang and
                  Lu Wang and
                  Weizhu Chen},
  title        = {LoRA: Low-Rank Adaptation of Large Language Models},
  booktitle    = {The Tenth International Conference on Learning Representations, {ICLR}
                  2022, Virtual Event, April 25-29, 2022},
  publisher    = {OpenReview.net},
  year         = {2022},
  url          = {https://openreview.net/forum?id=nZeVKeeFYf9},
  bibsource    = {dblp computer science bibliography, https://dblp.org}
}

@article{DBLP:journals/tacl/WangGZZLLT21,
  author       = {Xiaozhi Wang and
                  Tianyu Gao and
                  Zhaocheng Zhu and
                  Zhengyan Zhang and
                  Zhiyuan Liu and
                  Juanzi Li and
                  Jian Tang},
  title        = {{KEPLER:} {A} Unified Model for Knowledge Embedding and Pre-trained
                  Language Representation},
  journal      = {Trans. Assoc. Comput. Linguistics},
  volume       = {9},
  pages        = {176--194},
  year         = {2021},
  url          = {https://doi.org/10.1162/tacl\_a\_00360},
  doi          = {10.1162/TACL\_A\_00360},
  bibsource    = {dblp computer science bibliography, https://dblp.org}
}

@article{DBLP:journals/corr/abs-2001-08361,
  author       = {Jared Kaplan and
                  Sam McCandlish and
                  Tom Henighan and
                  Tom B. Brown and
                  Benjamin Chess and
                  Rewon Child and
                  Scott Gray and
                  Alec Radford and
                  Jeffrey Wu and
                  Dario Amodei},
  title        = {Scaling Laws for Neural Language Models},
  journal      = {CoRR},
  volume       = {abs/2001.08361},
  year         = {2020},
  url          = {https://arxiv.org/abs/2001.08361},
  eprinttype   = {arXiv},
  eprint       = {2001.08361},
  bibsource    = {dblp computer science bibliography, https://dblp.org}
}

@inproceedings{DBLP:conf/emnlp/UstunBBNR22,
  author       = {Ahmet {\"{U}}st{\"{u}}n and
                  Arianna Bisazza and
                  Gosse Bouma and
                  Gertjan van Noord and
                  Sebastian Ruder},
  editor       = {Yoav Goldberg and
                  Zornitsa Kozareva and
                  Yue Zhang},
  title        = {Hyper-X: {A} Unified Hypernetwork for Multi-Task Multilingual Transfer},
  booktitle    = {Proceedings of the 2022 Conference on Empirical Methods in Natural
                  Language Processing, {EMNLP} 2022, Abu Dhabi, United Arab Emirates,
                  December 7-11, 2022},
  pages        = {7934--7949},
  publisher    = {Association for Computational Linguistics},
  year         = {2022},
  url          = {https://doi.org/10.18653/v1/2022.emnlp-main.541},
  doi          = {10.18653/V1/2022.EMNLP-MAIN.541},
  bibsource    = {dblp computer science bibliography, https://dblp.org}
}

@inproceedings{DBLP:conf/icml/BethuneGBGCA25,
  author       = {Louis B{\'{e}}thune and
                  David Grangier and
                  Dan Busbridge and
                  Eleonora Gualdoni and
                  Marco Cuturi and
                  Pierre Ablin},
  editor       = {Aarti Singh and
                  Maryam Fazel and
                  Daniel Hsu and
                  Simon Lacoste{-}Julien and
                  Felix Berkenkamp and
                  Tegan Maharaj and
                  Kiri Wagstaff and
                  Jerry Zhu},
  title        = {Scaling Laws for Forgetting during Finetuning with Pretraining Data
                  Injection},
  booktitle    = {Forty-second International Conference on Machine Learning, {ICML}
                  2025, Vancouver, BC, Canada, July 13-19, 2025},
  series       = {Proceedings of Machine Learning Research},
  publisher    = {{PMLR} / OpenReview.net},
  year         = {2025},
  url          = {https://proceedings.mlr.press/v267/bethune25a.html},
  bibsource    = {dblp computer science bibliography, https://dblp.org}
}

@inproceedings{DBLP:conf/nips/BrownMRSKDNSSAA20,
  author       = {Tom B. Brown and
                  Benjamin Mann and
                  Nick Ryder and
                  Melanie Subbiah and
                  Jared Kaplan and
                  Prafulla Dhariwal and
                  Arvind Neelakantan and
                  Pranav Shyam and
                  Girish Sastry and
                  Amanda Askell and
                  Sandhini Agarwal and
                  Ariel Herbert{-}Voss and
                  Gretchen Krueger and
                  Tom Henighan and
                  Rewon Child and
                  Aditya Ramesh and
                  Daniel M. Ziegler and
                  Jeffrey Wu and
                  Clemens Winter and
                  Christopher Hesse and
                  Mark Chen and
                  Eric Sigler and
                  Mateusz Litwin and
                  Scott Gray and
                  Benjamin Chess and
                  Jack Clark and
                  Christopher Berner and
                  Sam McCandlish and
                  Alec Radford and
                  Ilya Sutskever and
                  Dario Amodei},
  editor       = {Hugo Larochelle and
                  Marc'Aurelio Ranzato and
                  Raia Hadsell and
                  Maria{-}Florina Balcan and
                  Hsuan{-}Tien Lin},
  title        = {Language Models are Few-Shot Learners},
  booktitle    = {Advances in Neural Information Processing Systems 33: Annual Conference
                  on Neural Information Processing Systems 2020, NeurIPS 2020, December
                  6-12, 2020, virtual},
  year         = {2020},
  url          = {https://proceedings.neurips.cc/paper/2020/hash/1457c0d6bfcb4967418bfb8ac142f64a-Abstract.html},
  bibsource    = {dblp computer science bibliography, https://dblp.org}
}

@inproceedings{DBLP:conf/nips/Ouyang0JAWMZASR22,
  author       = {Long Ouyang and
                  Jeffrey Wu and
                  Xu Jiang and
                  Diogo Almeida and
                  Carroll L. Wainwright and
                  Pamela Mishkin and
                  Chong Zhang and
                  Sandhini Agarwal and
                  Katarina Slama and
                  Alex Ray and
                  John Schulman and
                  Jacob Hilton and
                  Fraser Kelton and
                  Luke Miller and
                  Maddie Simens and
                  Amanda Askell and
                  Peter Welinder and
                  Paul F. Christiano and
                  Jan Leike and
                  Ryan Lowe},
  editor       = {Sanmi Koyejo and
                  S. Mohamed and
                  A. Agarwal and
                  Danielle Belgrave and
                  K. Cho and
                  A. Oh},
  title        = {Training language models to follow instructions with human feedback},
  booktitle    = {Advances in Neural Information Processing Systems 35: Annual Conference
                  on Neural Information Processing Systems 2022, NeurIPS 2022, New Orleans,
                  LA, USA, November 28 - December 9, 2022},
  year         = {2022},
  url          = {http://papers.nips.cc/paper\_files/paper/2022/hash/b1efde53be364a73914f58805a001731-Abstract-Conference.html},
  bibsource    = {dblp computer science bibliography, https://dblp.org}
}

@inproceedings{DBLP:conf/iclr/GuY25,
  author       = {Jeffrey Gu and
                  Serena Yeung{-}Levy},
  title        = {Foundation Models Secretly Understand Neural Network Weights: Enhancing
                  Hypernetwork Architectures with Foundation Models},
  booktitle    = {The Thirteenth International Conference on Learning Representations,
                  {ICLR} 2025, Singapore, April 24-28, 2025},
  publisher    = {OpenReview.net},
  year         = {2025},
  url          = {https://openreview.net/forum?id=cADpvQgnqg},
  bibsource    = {dblp computer science bibliography, https://dblp.org}
}

@article{DBLP:journals/ijon/SuALPBL24,
  author       = {Jianlin Su and
                  Murtadha H. M. Ahmed and
                  Yu Lu and
                  Shengfeng Pan and
                  Wen Bo and
                  Yunfeng Liu},
  title        = {RoFormer: Enhanced transformer with Rotary Position Embedding},
  journal      = {Neurocomputing},
  volume       = {568},
  pages        = {127063},
  year         = {2024},
  url          = {https://doi.org/10.1016/j.neucom.2023.127063},
  doi          = {10.1016/J.NEUCOM.2023.127063},
  bibsource    = {dblp computer science bibliography, https://dblp.org}
}

@article{DBLP:journals/corr/abs-2412-15115,
  author       = {An Yang and
                  Baosong Yang and
                  Beichen Zhang and
                  Binyuan Hui and
                  Bo Zheng and
                  Bowen Yu and
                  Chengyuan Li and
                  Dayiheng Liu and
                  Fei Huang and
                  Haoran Wei and
                  Huan Lin and
                  Jian Yang and
                  Jianhong Tu and
                  Jianwei Zhang and
                  Jianxin Yang and
                  Jiaxi Yang and
                  Jingren Zhou and
                  Junyang Lin and
                  Kai Dang and
                  Keming Lu and
                  Keqin Bao and
                  Kexin Yang and
                  Le Yu and
                  Mei Li and
                  Mingfeng Xue and
                  Pei Zhang and
                  Qin Zhu and
                  Rui Men and
                  Runji Lin and
                  Tianhao Li and
                  Tingyu Xia and
                  Xingzhang Ren and
                  Xuancheng Ren and
                  Yang Fan and
                  Yang Su and
                  Yichang Zhang and
                  Yu Wan and
                  Yuqiong Liu and
                  Zeyu Cui and
                  Zhenru Zhang and
                  Zihan Qiu},
  title        = {Qwen2.5 Technical Report},
  journal      = {CoRR},
  volume       = {abs/2412.15115},
  year         = {2024},
  url          = {https://doi.org/10.48550/arXiv.2412.15115},
  doi          = {10.48550/ARXIV.2412.15115},
  eprinttype   = {arXiv},
  eprint       = {2412.15115},
  bibsource    = {dblp computer science bibliography, https://dblp.org}
}

@inproceedings{DBLP:conf/iclr/LoshchilovH19,
  author       = {Ilya Loshchilov and
                  Frank Hutter},
  title        = {Decoupled Weight Decay Regularization},
  booktitle    = {7th International Conference on Learning Representations, {ICLR} 2019,
                  New Orleans, LA, USA, May 6-9, 2019},
  publisher    = {OpenReview.net},
  year         = {2019},
  url          = {https://openreview.net/forum?id=Bkg6RiCqY7},
  bibsource    = {dblp computer science bibliography, https://dblp.org}
}

@inproceedings{DBLP:conf/iclr/KothaSR24,
  author       = {Suhas Kotha and
                  Jacob Mitchell Springer and
                  Aditi Raghunathan},
  title        = {Understanding Catastrophic Forgetting in Language Models via Implicit
                  Inference},
  booktitle    = {The Twelfth International Conference on Learning Representations,
                  {ICLR} 2024, Vienna, Austria, May 7-11, 2024},
  publisher    = {OpenReview.net},
  year         = {2024},
  url          = {https://openreview.net/forum?id=VrHiF2hsrm},
  bibsource    = {dblp computer science bibliography, https://dblp.org}
}

@inproceedings{DBLP:conf/emnlp/Li0FT24,
  author       = {Hongyu Li and
                  Liang Ding and
                  Meng Fang and
                  Dacheng Tao},
  editor       = {Yaser Al{-}Onaizan and
                  Mohit Bansal and
                  Yun{-}Nung Chen},
  title        = {Revisiting Catastrophic Forgetting in Large Language Model Tuning},
  booktitle    = {Findings of the Association for Computational Linguistics: {EMNLP}
                  2024, Miami, Florida, USA, November 12-16, 2024},
  series       = {Findings of {ACL}},
  pages        = {4297--4308},
  publisher    = {Association for Computational Linguistics},
  year         = {2024},
  url          = {https://doi.org/10.18653/v1/2024.findings-emnlp.249},
  doi          = {10.18653/V1/2024.FINDINGS-EMNLP.249},
  bibsource    = {dblp computer science bibliography, https://dblp.org}
}

@inproceedings{DBLP:conf/naacl/HajipourYSF24,
  author       = {Hossein Hajipour and
                  Ning Yu and
                  Cristian{-}Alexandru Staicu and
                  Mario Fritz},
  editor       = {Kevin Duh and
                  Helena G{\'{o}}mez{-}Adorno and
                  Steven Bethard},
  title        = {SimSCOOD: Systematic Analysis of Out-of-Distribution Generalization
                  in Fine-tuned Source Code Models},
  booktitle    = {Findings of the Association for Computational Linguistics: {NAACL}
                  2024, Mexico City, Mexico, June 16-21, 2024},
  series       = {Findings of {ACL}},
  pages        = {1400--1416},
  publisher    = {Association for Computational Linguistics},
  year         = {2024},
  url          = {https://doi.org/10.18653/v1/2024.findings-naacl.90},
  doi          = {10.18653/V1/2024.FINDINGS-NAACL.90},
  bibsource    = {dblp computer science bibliography, https://dblp.org}
}

@article{DBLP:journals/corr/abs-2303-08774,
  author       = {OpenAI},
  title        = {{GPT-4} Technical Report},
  journal      = {CoRR},
  volume       = {abs/2303.08774},
  year         = {2023},
  url          = {https://doi.org/10.48550/arXiv.2303.08774},
  doi          = {10.48550/ARXIV.2303.08774},
  eprinttype   = {arXiv},
  eprint       = {2303.08774},
  bibsource    = {dblp computer science bibliography, https://dblp.org}
}

@article{DBLP:journals/corr/abs-2505-12345,
  author       = {Qizhou Chen and
                  Dakan Wang and
                  Taolin Zhang and
                  Zaoming Yan and
                  Chengsong You and
                  Chengyu Wang and
                  Xiaofeng He},
  title        = {UniEdit: {A} Unified Knowledge Editing Benchmark for Large Language
                  Models},
  journal      = {CoRR},
  volume       = {abs/2505.12345},
  year         = {2025},
  url          = {https://doi.org/10.48550/arXiv.2505.12345},
  doi          = {10.48550/ARXIV.2505.12345},
  eprinttype   = {arXiv},
  eprint       = {2505.12345},
  bibsource    = {dblp computer science bibliography, https://dblp.org}
}

@inproceedings{DBLP:conf/conll/LevySCZ17,
  author       = {Omer Levy and
                  Minjoon Seo and
                  Eunsol Choi and
                  Luke Zettlemoyer},
  editor       = {Roger Levy and
                  Lucia Specia},
  title        = {Zero-Shot Relation Extraction via Reading Comprehension},
  booktitle    = {Proceedings of the 21st Conference on Computational Natural Language
                  Learning (CoNLL 2017), Vancouver, Canada, August 3-4, 2017},
  pages        = {333--342},
  publisher    = {Association for Computational Linguistics},
  year         = {2017},
  url          = {https://doi.org/10.18653/v1/K17-1034},
  doi          = {10.18653/V1/K17-1034},
  bibsource    = {dblp computer science bibliography, https://dblp.org}
}

@inproceedings{DBLP:conf/icml/ThedeRBAH25,
  author       = {Lukas Thede and
                  Karsten Roth and
                  Matthias Bethge and
                  Zeynep Akata and
                  Thomas Hartvigsen},
  editor       = {Aarti Singh and
                  Maryam Fazel and
                  Daniel Hsu and
                  Simon Lacoste{-}Julien and
                  Felix Berkenkamp and
                  Tegan Maharaj and
                  Kiri Wagstaff and
                  Jerry Zhu},
  title        = {WikiBigEdit: Understanding the Limits of Lifelong Knowledge Editing
                  in LLMs},
  booktitle    = {Forty-second International Conference on Machine Learning, {ICML}
                  2025, Vancouver, BC, Canada, July 13-19, 2025},
  series       = {Proceedings of Machine Learning Research},
  publisher    = {{PMLR} / OpenReview.net},
  year         = {2025},
  url          = {https://proceedings.mlr.press/v267/thede25a.html},
  bibsource    = {dblp computer science bibliography, https://dblp.org}
}

@article{DBLP:journals/corr/abs-2306-06031,
  author       = {Hongyang Yang and
                  Xiao{-}Yang Liu and
                  Christina Dan Wang},
  title        = {FinGPT: Open-Source Financial Large Language Models},
  journal      = {CoRR},
  volume       = {abs/2306.06031},
  year         = {2023},
  url          = {https://doi.org/10.48550/arXiv.2306.06031},
  doi          = {10.48550/ARXIV.2306.06031},
  eprinttype   = {arXiv},
  eprint       = {2306.06031},
  bibsource    = {dblp computer science bibliography, https://dblp.org}
}

@article{DBLP:journals/csur/LingZLDZWCLCZZPMPCWLCCW26,
  author       = {Chen Ling and
                  Xujiang Zhao and
                  Jiaying Lu and
                  Chengyuan Deng and
                  Can Zheng and
                  Junxiang Wang and
                  Tanmoy Chowdhury and
                  Yun Li and
                  Hejie Cui and
                  Xuchao Zhang and
                  Tianjiao Zhao and
                  Amit Panalkar and
                  Dhagash Mehta and
                  Stefano Pasquali and
                  Wei Cheng and
                  Haoyu Wang and
                  Yanchi Liu and
                  Zhengzhang Chen and
                  Haifeng Chen and
                  Chris White and
                  Quanquan Gu and
                  Jian Pei and
                  Carl Yang and
                  Liang Zhao},
  title        = {Domain Specialization as the Key to Make Large Language Models Disruptive:
                  {A} Comprehensive Survey},
  journal      = {{ACM} Comput. Surv.},
  volume       = {58},
  number       = {3},
  pages        = {79:1--79:39},
  year         = {2026},
  url          = {https://doi.org/10.1145/3764579},
  doi          = {10.1145/3764579},
  bibsource    = {dblp computer science bibliography, https://dblp.org}
}

@inproceedings{DBLP:conf/emnlp/GekhmanYAEFRH24,
  author       = {Zorik Gekhman and
                  Gal Yona and
                  Roee Aharoni and
                  Matan Eyal and
                  Amir Feder and
                  Roi Reichart and
                  Jonathan Herzig},
  editor       = {Yaser Al{-}Onaizan and
                  Mohit Bansal and
                  Yun{-}Nung Chen},
  title        = {Does Fine-Tuning LLMs on New Knowledge Encourage Hallucinations?},
  booktitle    = {Proceedings of the 2024 Conference on Empirical Methods in Natural
                  Language Processing, {EMNLP} 2024, Miami, FL, USA, November 12-16,
                  2024},
  pages        = {7765--7784},
  publisher    = {Association for Computational Linguistics},
  year         = {2024},
  url          = {https://doi.org/10.18653/v1/2024.emnlp-main.444},
  doi          = {10.18653/V1/2024.EMNLP-MAIN.444},
  bibsource    = {dblp computer science bibliography, https://dblp.org}
}

@article{DBLP:journals/corr/abs-2501-13687,
  author       = {Sara Kothari and
                  Ayush Gupta},
  title        = {Question Answering on Patient Medical Records with Private Fine-Tuned
                  LLMs},
  journal      = {CoRR},
  volume       = {abs/2501.13687},
  year         = {2025},
  url          = {https://doi.org/10.48550/arXiv.2501.13687},
  doi          = {10.48550/ARXIV.2501.13687},
  eprinttype   = {arXiv},
  eprint       = {2501.13687},
  bibsource    = {dblp computer science bibliography, https://dblp.org}
}

@article{DBLP:journals/corr/abs-2510-11588,
  author       = {Jiateng Liu and
                  Zhenhailong Wang and
                  Xiaojiang Huang and
                  Yingjie Li and
                  Xing Fan and
                  Xiang Li and
                  Chenlei Guo and
                  Ruhi Sarikaya and
                  Heng Ji},
  title        = {Analyzing and Internalizing Complex Policy Documents for {LLM} Agents},
  journal      = {CoRR},
  volume       = {abs/2510.11588},
  year         = {2025},
  url          = {https://doi.org/10.48550/arXiv.2510.11588},
  doi          = {10.48550/ARXIV.2510.11588},
  eprinttype   = {arXiv},
  eprint       = {2510.11588},
  bibsource    = {dblp computer science bibliography, https://dblp.org}
}

@inproceedings{DBLP:conf/nips/Wei0SBIXCLZ22,
  author       = {Jason Wei and
                  Xuezhi Wang and
                  Dale Schuurmans and
                  Maarten Bosma and
                  Brian Ichter and
                  Fei Xia and
                  Ed H. Chi and
                  Quoc V. Le and
                  Denny Zhou},
  editor       = {Sanmi Koyejo and
                  S. Mohamed and
                  A. Agarwal and
                  Danielle Belgrave and
                  K. Cho and
                  A. Oh},
  title        = {Chain-of-Thought Prompting Elicits Reasoning in Large Language Models},
  booktitle    = {Advances in Neural Information Processing Systems 35: Annual Conference
                  on Neural Information Processing Systems 2022, NeurIPS 2022, New Orleans,
                  LA, USA, November 28 - December 9, 2022},
  year         = {2022},
  url          = {http://papers.nips.cc/paper\_files/paper/2022/hash/9d5609613524ecf4f15af0f7b31abca4-Abstract-Conference.html},
  bibsource    = {dblp computer science bibliography, https://dblp.org}
}

@inproceedings{DBLP:conf/aaai/LiuZ0WJD020,
  author       = {Weijie Liu and
                  Peng Zhou and
                  Zhe Zhao and
                  Zhiruo Wang and
                  Qi Ju and
                  Haotang Deng and
                  Ping Wang},
  title        = {{K-BERT:} Enabling Language Representation with Knowledge Graph},
  booktitle    = {The Thirty-Fourth {AAAI} Conference on Artificial Intelligence, {AAAI}
                  2020, The Thirty-Second Innovative Applications of Artificial Intelligence
                  Conference, {IAAI} 2020, The Tenth {AAAI} Symposium on Educational
                  Advances in Artificial Intelligence, {EAAI} 2020, New York, NY, USA,
                  February 7-12, 2020},
  pages        = {2901--2908},
  publisher    = {{AAAI} Press},
  year         = {2020},
  url          = {https://doi.org/10.1609/aaai.v34i03.5681},
  doi          = {10.1609/AAAI.V34I03.5681},
  bibsource    = {dblp computer science bibliography, https://dblp.org}
}

@inproceedings{DBLP:conf/acl/WangTDWHJCJZ21,
  author       = {Ruize Wang and
                  Duyu Tang and
                  Nan Duan and
                  Zhongyu Wei and
                  Xuanjing Huang and
                  Jianshu Ji and
                  Guihong Cao and
                  Daxin Jiang and
                  Ming Zhou},
  editor       = {Chengqing Zong and
                  Fei Xia and
                  Wenjie Li and
                  Roberto Navigli},
  title        = {K-Adapter: Infusing Knowledge into Pre-Trained Models with Adapters},
  booktitle    = {Findings of the Association for Computational Linguistics: {ACL/IJCNLP}
                  2021, Online Event, August 1-6, 2021},
  series       = {Findings of {ACL}},
  volume       = {{ACL-IJCNLP} 2021},
  pages        = {1405--1418},
  publisher    = {Association for Computational Linguistics},
  year         = {2021},
  url          = {https://doi.org/10.18653/v1/2021.findings-acl.121},
  doi          = {10.18653/V1/2021.FINDINGS-ACL.121},
  bibsource    = {dblp computer science bibliography, https://dblp.org}
}

@inproceedings{DBLP:conf/acl/ZhangHLJSL19,
  author       = {Zhengyan Zhang and
                  Xu Han and
                  Zhiyuan Liu and
                  Xin Jiang and
                  Maosong Sun and
                  Qun Liu},
  editor       = {Anna Korhonen and
                  David R. Traum and
                  Llu{\'{\i}}s M{\`{a}}rquez},
  title        = {{ERNIE:} Enhanced Language Representation with Informative Entities},
  booktitle    = {Proceedings of the 57th Conference of the Association for Computational
                  Linguistics, {ACL} 2019, Florence, Italy, July 28- August 2, 2019,
                  Volume 1: Long Papers},
  pages        = {1441--1451},
  publisher    = {Association for Computational Linguistics},
  year         = {2019},
  url          = {https://doi.org/10.18653/v1/p19-1139},
  doi          = {10.18653/V1/P19-1139},
  bibsource    = {dblp computer science bibliography, https://dblp.org}
}

@inproceedings{DBLP:conf/emnlp/OvadiaBME24,
  author       = {Oded Ovadia and
                  Menachem Brief and
                  Moshik Mishaeli and
                  Oren Elisha},
  editor       = {Yaser Al{-}Onaizan and
                  Mohit Bansal and
                  Yun{-}Nung Chen},
  title        = {Fine-Tuning or Retrieval? Comparing Knowledge Injection in {LLMs}},
  booktitle    = {Proceedings of the 2024 Conference on Empirical Methods in Natural
                  Language Processing, {EMNLP} 2024, Miami, FL, USA, November 12-16,
                  2024},
  pages        = {237--250},
  publisher    = {Association for Computational Linguistics},
  year         = {2024},
  url          = {https://aclanthology.org/2024.emnlp-main.15},
  bibsource    = {dblp computer science bibliography, https://dblp.org}
}
\bibliographystyle{plainnat}

%%%%%%%%%%%%%%%%%%%%%%%%%%%%%%%%%%%%%%%%%%%%%%%%%%%%%%%%%%%%

\appendix

\section{Calculating Training Compute for Hypernetworks} \label{sec:training_flops}

\citet{DBLP:journals/corr/abs-2001-08361} uses $2N$ as the number of floating-point operations (FLOPs) required for a transformer forward pass per token, and $6N$ per training iteration per token, where $N$ is the number of non-embedding parameters. In our experiments, we consider hypernetworks that contain a comparable number of parameters to that of the target model. In order to compute the number of FLOPs during training with hypernetworks, we adapt the formula appropriately:
\begin{enumerate}
    \item \textbf{Cost of hypernetwork forward pass:} $2 n_h N_h$ where $N_h$ is the number of non-embedding hypernetwork parameters and $n_h$ is the number of input tokens in the hypernetwork.
    \item \textbf{Cost of target model forward pass:} $2 n_t N_t$ where $N_t$ is the number of non-embedding parameters in the target model and $n_t$ is the number of input tokens in the target model.
    \item \textbf{Cost of target model backward pass:} $2 n_t N_t \frac{L_{target}}{L}$ where $L$ is the number of layers in the target model and $L_{target}$ is the number of layers targeted by the hypernetwork, assuming we target the last $L_{target}$ layers. Note that in order to compute the gradients in the hypernetwork, we must backpropagate the gradients through the last $L_{target}$ layers of the target network (and no further). During this backward pass, we only need to compute the gradients of the loss function with respect to the previous layer's activations, and we do not need to compute the gradients with respect to the weights in the target network. As a result, for any linear layer, we only need one matrix multiplication in the backward pass, hence the coefficient $2$.
    \item \textbf{Cost of hypernetwork backward pass:} $4 n_h N_h$.
\end{enumerate}

Thus the total cost $C$ of a training iteration in FLOPs on one example is:
\begin{equation}
    C = 6 n_h N_h + 2 n_t N_t \left( 1 + \frac{L_{target}}{L} \right).
\end{equation}

In our experiments, the average length of the queries for the target model is roughly 15 tokens, and so we fix $n_t = 15$.
Each injected fact contributes on average $11.4$ tokens to the hypernetwork
input, so we set $n_h = 11.4 \times N_{\text{facts}}$. For width, depth, and 
target-model scaling, which use $N_{\text{facts}} = 4$ injected facts, 
which gives $n_h = 45.6$.

Tables~\ref{tab:flops_width}--\ref{tab:flops_target} report the 
per-example FLOPs for each scaling axis in our experiments. For the depth scaling 
experiments, varying $L_{\text{HN}}$ from 1 to 16 increases $N_h$ 
by only 1.9\%, since the LoRA projection heads dominate $N_h$ at 
$d_{\text{model}}=256$. Consequently, depth scaling is effectively 
\textbf{iso-compute}, with FLOPs per example varying by less than 2\% 
across the entire depth range, meaning that performance differences 
observed in Section~\ref{sec:depth_scaling} are attributable to 
architectural capacity rather than additional compute. Width and 
fact-count scaling are not iso-compute: varying $d_{\text{model}}$ 
from 64 to 1024 introduces a $7.4\times$ compute range, and varying 
$N_{\text{facts}}$ from 2 to 52 introduces a $13.7\times$ compute 
range through $n_h$.

\begin{table}[h]
  \caption{FLOPs per example for hypernetwork width scaling 
  (target: \texttt{Qwen2.5-1.5B}, $L_{\text{HN}}=4$, $n_h=45.6$).}
  \label{tab:flops_width}
  \centering
  \small
  \begin{tabular}{rrrrr}
    \toprule
    $d_{\text{model}}$ & $L_{\text{HN}}$ & 
    $N_h$ (w/ embed) & $N_h$ (w/o embed) & 
    FLOPs/example \\
    \midrule
    64   & 4 & 167,398,376   & 157,691,816   & $1.021 \times 10^{11}$ \\
    128  & 4 & 323,068,776   & 313,362,216   & $1.447 \times 10^{11}$ \\
    256  & 4 & 635,580,904   & 625,874,344   & $2.302 \times 10^{11}$ \\
    512  & 4 & 1,265,323,752 & 1,255,617,192 & $4.025 \times 10^{11}$ \\
    1024 & 4 & 2,543,683,816 & 2,533,977,256 & $7.523 \times 10^{11}$ \\
    \bottomrule
  \end{tabular}
\end{table}

\begin{table}[h]
  \caption{FLOPs per example for hypernetwork depth scaling 
  (target: \texttt{Qwen2.5-1.5B}, $d_{\text{model}}=256$, $n_h=45.6$).}
  \label{tab:flops_depth}
  \centering
  \small
  \begin{tabular}{rrrrr}
    \toprule
    $d_{\text{model}}$ & $L_{\text{HN}}$ & 
    $N_h$ (w/ embed) & $N_h$ (w/o embed) & 
    FLOPs/example \\
    \midrule
    256 & 1  & 633,211,624 & 623,505,064 & $2.296 \times 10^{11}$ \\
    256 & 2  & 634,001,384 & 624,294,824 & $2.298 \times 10^{11}$ \\
    256 & 4  & 635,580,904 & 625,874,344 & $2.302 \times 10^{11}$ \\
    256 & 8  & 638,739,944 & 629,033,384 & $2.311 \times 10^{11}$ \\
    256 & 16 & 645,058,024 & 635,351,464 & $2.328 \times 10^{11}$ \\
    \bottomrule
  \end{tabular}
\end{table}

\begin{table}[h]
  \caption{FLOPs per example for fact-count scaling 
  (target: \texttt{Qwen2.5-1.5B}, $d_{\text{model}}=256$, 
  $L_{\text{HN}}=4$, fact embedding dim $=512$). The 52-fact run uses 
  $n_h=512$ due to the hypernetwork context window limit.}
  \label{tab:flops_facts}
  \centering
  \small
  \begin{tabular}{rrrrr}
    \toprule
    $N_{\text{facts}}$ & $n_h$ & 
    $N_h$ (w/ embed) & $N_h$ (w/o embed) & 
    FLOPs/example \\
    \midrule
    2  & 22.8  & 703,641,512 & 625,989,032 & $1.446 \times 10^{11}$ \\
    4  & 45.6  & 703,641,512 & 625,989,032 & $2.302 \times 10^{11}$ \\
    8  & 91.2  & 703,641,512 & 625,989,032 & $4.014 \times 10^{11}$ \\
    16 & 182.4 & 703,641,512 & 625,989,032 & $7.440 \times 10^{11}$ \\
    32 & 364.8 & 703,641,512 & 625,989,032 & $1.429 \times 10^{12}$ \\
    52 & 512   & 703,641,512 & 625,989,032 & $1.982 \times 10^{12}$ \\
    \bottomrule
  \end{tabular}
\end{table}

\begin{table}[h]
  \caption{FLOPs per example for target model scaling 
  ($d_{\text{model}}=256$, $L_{\text{HN}}=4$, $n_h=45.6$).}
  \label{tab:flops_target}
  \centering
  \small
  \begin{tabular}{lrrr}
    \toprule
    Target & $N_h$ (w/ embed) & $N_h$ (w/o embed) & 
    FLOPs/example \\
    \midrule
    0.5B & 309,772,752   & 300,066,192   & $9.820 \times 10^{10}$ \\
    1.5B & 635,580,904   & 625,874,344   & $2.302 \times 10^{11}$ \\
    3B   & 909,808,128   & 900,101,568   & $3.665 \times 10^{11}$ \\
    7B   & 2,100,000,000 & 2,090,293,440 & $8.795 \times 10^{11}$ \\
    14B  & 2,800,000,000 & 2,790,293,440 & $1.292 \times 10^{12}$ \\
    \bottomrule
  \end{tabular}
\end{table}

\section{Domain Classification}
\label{sec:domain}

In this section, we provide further details on the domain classification 
procedure, which is utilized in the construction of our dataset, 
\texttt{MegaWikiQA}. Each example in \texttt{MegaWikiQA} is derived from 
a triplet $(s, r, o)$ in \texttt{Wikidata5M}, and we assign each triplet 
to one of 39 semantic domains (e.g., geography, political science, music) 
using a two-stage classification pipeline described below. We assign each 
example to a semantic domain to enable controlled evaluation of 
generalization across domains. Domain labels allow us to construct OOD 
splits by holding out entire domains during training, providing a stricter 
test of whether injected knowledge generalizes beyond the distribution of 
observed facts.

\paragraph{Domain taxonomy.}
We adopt a fixed set of 39 domains, some were extracted from prior work on knowledge editing
benchmarks (e.g., \texttt{UniEdit}~\citep{DBLP:journals/corr/abs-2505-12345}), ensuring consistency with existing datasets and
facilitating comparability. To align our data with this taxonomy, we train a
domain classifier on triplet-domain pairs derived from the \texttt{UniEdit} dataset and GPT-4.1.

\paragraph{Classification pipeline.}
Given a triplet $(s, r, o)$, we assign a domain label using a two-stage
pipeline. In the first stage, we prompt GPT-4.1 to predict a domain label from
the predefined set of 39 domains based on the semantics of the triplet. In the
second stage, we apply a fine-tuned LLM with a classifier head to validate the
prediction. For training data, we first collected around 100K high quality labelled samples using multiple calls of GPT models \& voting ensemble. We then performed supervised finetuning of classifier with class-wise cross-entropy loss. We perform rejection filtering by discarding examples for which the
classifier assigns low confidence, retaining only high-confidence domain labels.

\paragraph{Quality control.}
This process yields domain labels with over 95\% manually verified classification accuracy on a
held-out validation set, and a 92\% agreement rate with GPT-4.1 predictions.
Filtering low-confidence predictions reduces label noise and ensures reliable
domain assignments for downstream evaluation.

\paragraph{Application to multi-hop examples.}
Domain labels are first assigned to single-hop triplets and then propagated to
multi-hop examples based on their underlying triplet sequences. Using these labels,
we construct stratified splits that are balanced in both domain coverage and reasoning
complexity (i.e., number of hops), ensuring that scaling experiments are not confounded by
distributional imbalances.

\begin{figure}[h]
    \centering
    \includegraphics[width=\linewidth]{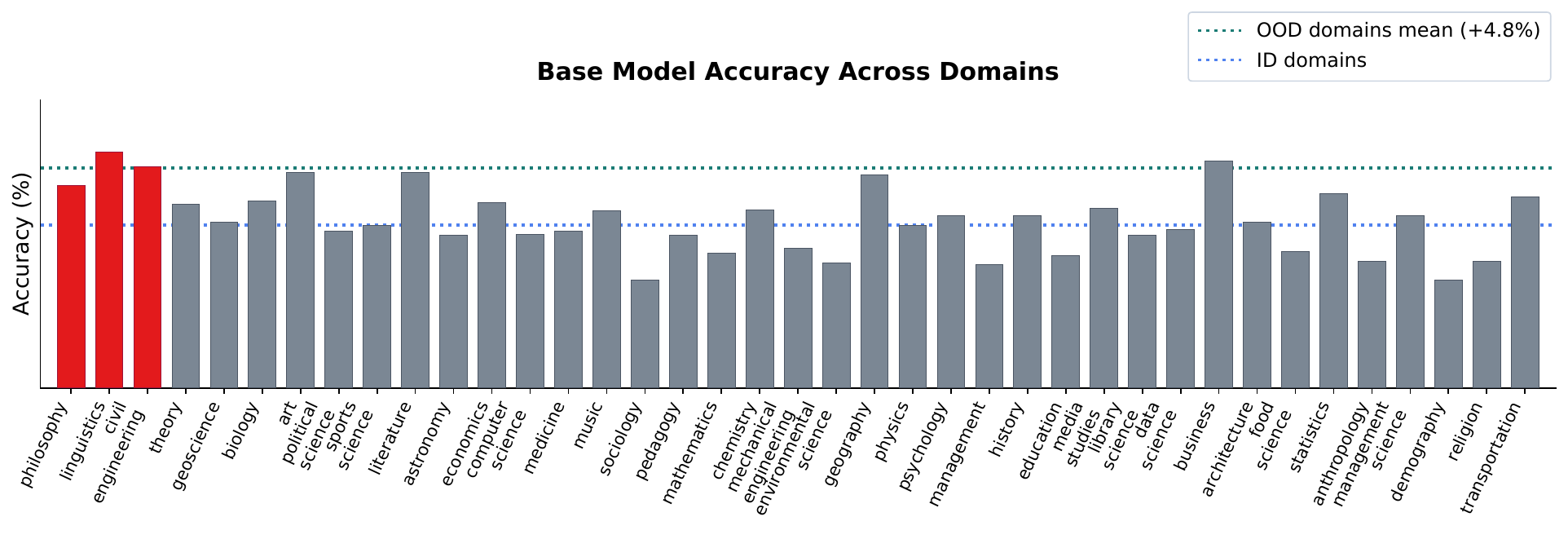}
    \caption{Base model accuracy across all 39 knowledge domains. 
    The three held-out OOD domains (\textit{philosophy}, 
    \textit{linguistics}, \textit{civil engineering}) exhibit high base 
    model accuracy, lying approximately 4.8\% above the mean of the 
    in-distribution domains.}
    \label{fig:domain_accuracy}
\end{figure}

\paragraph{Domain-level base model accuracy.}
Figure~\ref{fig:domain_accuracy} shows the base model accuracy across all 
39 domains. The three held-out OOD domains (\textit{philosophy}, 
\textit{linguistics}, and \textit{civil engineering}) exhibit notably 
higher base model accuracy than the in-distribution domains, lying 
approximately 4.8\% above the in-distribution mean. This has an important 
implication for the interpretation of our OOD results: because the frozen 
target model already has stronger prior knowledge of these domains, the OOD non-rephrased loss is close to the ID validation loss across all 
scaling experiments, as observed in the scaling law figures in 
Section~\ref{sec:experiments}. The OOD rephrased 
and OOD MCQ splits, which test robustness to linguistic variation and 
format shift respectively, provide a stricter evaluation that is less 
confounded by base model familiarity.

\section{Dataset Statistics}
\label{sec:dataset_stats}

In this section, we provide full statistics for the \texttt{MegaWikiQA} 
dataset. Table~\ref{tab:dataset} summarizes all splits used in our 
experiments. All evaluation splits are disjoint from the training set at 
the triplet level, and the three OOD domains (\textit{philosophy}, 
\textit{linguistics}, \textit{civil engineering}) are fully withheld 
during training.

\begin{table}[h]
  \caption{Dataset statistics. All splits have zero entity and triplet 
  overlap with the training set. OOD domains are fully held out during 
  training.}
  \label{tab:dataset}
  \centering
  \begin{tabular}{llrr}
    \toprule
    Split & Type & Number of hops & Examples \\
    \midrule
    Training          & In-distribution  & 1, 2, 3, 4 & 1,250,000 \\
    ID Eval           & In-distribution  & 1, 2, 3, 4 & 10,000    \\
    OOD Non-rephrased & Held-out domains & 1, 2, 3, 4 & 10,000    \\
    OOD Rephrased     & Held-out domains & 1, 2, 3, 4 & 10,000    \\
    OOD MCQ           & Held-out domains & 1, 2, 3, 4 & 10,000    \\
    \bottomrule
  \end{tabular}
\end{table}
%%%%%%%%%%%%%%%%%%%%%%%%%%%%%%%%%%%%%%%%%%%%%%%%%%%%%%%%%%%%

\section{Hypernetwork Architecture Details}
\label{sec:hypernet_details}

This section covers full architectural details of the hypernetwork 
$g_\phi$ introduced in Section~\ref{sec:hypernet_architecture}. This 
includes the input encoding scheme, the transformer encoder design, the 
LoRA weight generation procedure, the target layer selection strategy, 
and the hyperparameter tuning approach used across all scaling experiments.

\subsection{Input Encoding}
Each fact $f_i \in \mathcal{F}$ is serialized as a natural language sentence
and tokenized using the same tokenizer as the target language model, with a
maximum sequence length of 128 tokens. The fact set $\mathcal{F}$ is concatenated and
processed by the hypernetwork transformer encoder. To obtain a single fixed-size
representation of $\mathcal{F}$, we apply masked mean pooling over all
non-padding token positions.
%\begin{equation}
%    \mathbf{h} = \frac{\sum_{t} \mathbf{x}_t \cdot m_t}{\sum_{t} m_t}
%    \in \mathbb{R}^{d_{\text{model}}}
%\end{equation}
%where $\mathbf{x}_t$ is the contextualized hidden state at position $t$,
%$m_t = 0$ if the input token at $t$ is a padding token, and $m_t = 1$ otherwise.
This pooling strategy
is parameter-free and treats all token positions symmetrically, avoiding the
inductive bias of token-specific aggregation schemes such as \texttt{[CLS]}
pooling.

\subsection{Encoder Architecture}
The hypernetwork transformer encoder uses rotary positional embeddings
(RoPE)~\citet{DBLP:journals/ijon/SuALPBL24} applied to the query and key projections of each
attention layer, providing relative positional awareness without absolute
position bias. Each transformer block follows a post-norm architecture, where
layer normalization is applied after the residual connection rather than
before, which we find improves training stability across the range of
hypernetwork depths explored in our scaling experiments.

\subsection{LoRA Weight Generation}
For each target weight matrix $W \in \mathbb{R}^{d_{\text{out}} \times
d_{\text{in}}}$ in the selected layers of the target model, the hypernetwork
maintains two dedicated linear projection heads that map the pooled
representation $\mathbf{h}$ to the corresponding low-rank factors:
\begin{equation}
    A = \left(\tanh\!\left(W_A \mathbf{h}\right)\right)^{\top} \in \mathbb{R}^{r \times d_{\text{in}}}, \hspace{5em} B = \left(\tanh\!\left(W_B \mathbf{h}\right)\right)^{\top} \in \mathbb{R}^{d_{\text{out}} \times r},
\end{equation}
where $r$ is the LoRA rank, $W_A \in \mathbb{R}^{(r \cdot d_{\text{in}}) \times d_{\text{model}}}$ and
$W_B \in \mathbb{R}^{(d_{\text{out}} \cdot r) \times d_{\text{model}}}$ are
learned projection matrices, and the outputs are reshaped from vectors to
matrices of the indicated dimensions. The weight adaptation applied to the
target model is then $\Delta W = \alpha c BA / r$ where $r$ is the LoRA rank, $\alpha$ is the scaling factor, and $c = 0.01$
is a fixed scalar that controls the magnitude of the generated adaptations.
We use $r = 4$ and $\alpha = 8$ across all experiments.
LoRA adaptations are applied to all linear layers within the selected
transformer blocks of the target model, covering both attention projection
matrices and feed-forward layers, to maximize the expressivity of the
generated adaptations.

\subsection{Target Layer Selection}
We apply hypernetwork-generated LoRA adaptations to the later $\lfloor L/2
\rfloor$ layers of the target model, where $L$ denotes the total number of
transformer layers. This later-layer adaptation strategy concentrates the
injection signal into the higher-level semantic representations of the target
model, which prior work has shown to be the primary locus of factual
knowledge~\citep{DBLP:conf/nips/MengBAB22}. The proportion $\lfloor L/2
\rfloor$ is held fixed across all target model sizes, ensuring that the
fraction of adapted parameters scales consistently with model depth.

\subsection{Training Objective}
The hypernetwork is trained end-to-end by minimizing cross-entropy loss over the
answer tokens under the adapted target model. Gradients flow through
the target model and back into the hypernetwork parameters $\phi$
via the LoRA adaptation matrices. The target model parameters $\theta$ receive
no gradient updates and remain frozen throughout.

\subsection{Hyperparameter Tuning}
\begin{figure}[h]
    \centering
    \includegraphics[width=\linewidth]{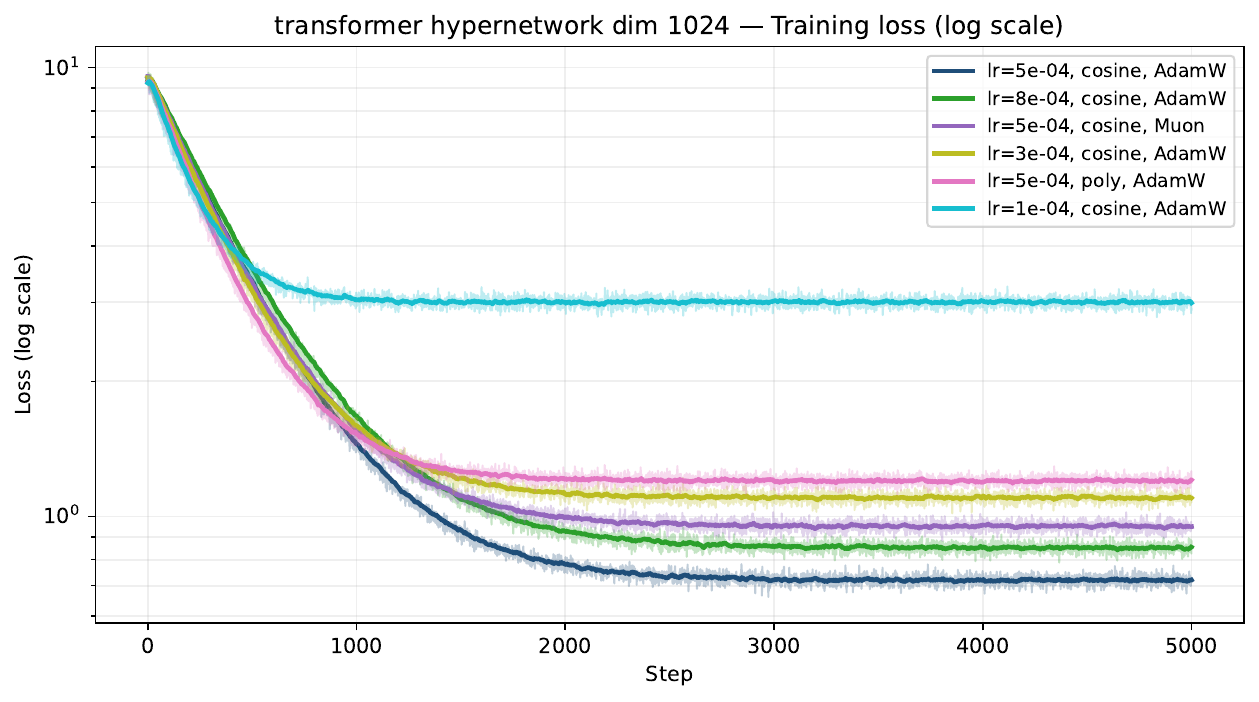}
    \caption{Hyperparameter sweep results for the largest width configuration 
    ($d_{\text{model}} = 1024$), showing training loss on a log scale across 
    different learning rates, schedulers, and optimizers on a small-scale 
    subset. Cosine annealing with AdamW with learning rate $5 \times 10^{-4}$ 
    achieves the lowest final training loss, motivating its selection for all 
    scaling experiments.}
    \label{fig:hp_sweep}
\end{figure}

We use the AdamW~\citep{DBLP:conf/iclr/LoshchilovH19} optimizer with weight 
decay $\lambda = 10^{-5}$ across all experiments. To identify the best 
scheduler and learning rate range efficiently, we first conducted a broad 
sweep across schedulers and learning rates on a small subset of 10K--80K 
training examples, before scaling up to the full 1.25M-example training set. 
Figure~\ref{fig:hp_sweep} shows an example of such a sweep for the largest 
width configuration ($d_{\text{model}} = 1024$), illustrating how cosine 
annealing with AdamW consistently outperformed polynomial decay and the Muon 
optimizer across all learning rates tested. Based on these small-scale 
experiments, we adopted cosine annealing with AdamW as the scheduler for all 
scaling experiments.

Rather than performing an exhaustive hyperparameter search for each 
configuration at full scale, we adopt an efficient \emph{anchor-and-interpolate} 
strategy for learning rate selection. For each scaling variable (e.g., 
hypernetwork depth, width, etc), we first run experiments at the lowest and highest 
values in the experimental range, followed by a third experiment at the 
midpoint (arithmetic mean). We perform hyperparameter optimization at these three anchor points, 
then fit a power-law of the form $\eta^*(x) = a \cdot x^b + c$, where $x$ is 
the scaling variable and the coefficients $a, b, c$ are estimated via 
least-squares regression in log scale. Learning rates for all intermediate 
configurations are obtained by evaluating this fitted curve. The optimal 
learning rate ranges observed across scaling axes were as follows: for width 
scaling, $\eta^*$ decreased from $5 \times 10^{-4}$ to $6 \times 10^{-5}$ 
as $d_{\text{model}}$ increased from 64 to 1024; for depth scaling, the 
range was narrower ($2 \times 10^{-4}$ to $9 \times 10^{-5}$) owing to the 
similar parameter counts across depth configurations; for target model 
scaling, $\eta^*$ ranged from $6 \times 10^{-4}$ to $4 \times 10^{-5}$; 
and for fact count scaling, from $2 \times 10^{-4}$ to $7 \times 10^{-5}$.

The number of training epochs is selected per experiment based on 
convergence behavior. We continue training until the rate of improvement becomes 
negligible and then continue training for a small number of additional epochs to 
confirm convergence. In practice, the number of epochs ranged from 5 to 8 across all 
experiments. For configurations with lower FLOPs per example (e.g., small 
hypernetwork widths or few injected facts), training is extended for additional epochs 
to ensure that these runs reach a comparable level of convergence to their 
higher-compute counterparts, enabling fair comparisons across the scaling 
axis. Batch size is selected independently for each experiment based on 
training loss stability.

\subsection{Target Model Scaling}

To characterize how knowledge injection performance scales with the capacity
of the frozen target model, we conduct experiments across five model sizes
from the \texttt{Qwen2.5} family~\citet{DBLP:journals/corr/abs-2412-15115}: \textbf{0.5B}, \textbf{1.5B}, \textbf{3B}, \textbf{7B}, and \textbf{14B}
parameters. The hypernetwork architecture is held fixed across all target
model sizes, and LoRA adaptations are applied to the later $\lfloor L/2
\rfloor$ layers in each case, where $L$ varies with target model depth.
This design isolates the effect of target model capacity from hypernetwork
capacity, enabling a clean comparison of the effect of increasing training compute in the target model versus the hypernetwork.

\section{LoRA Rank Scaling}
\label{sec:lora_rank_scaling}

\begin{figure}[h]
    \centering
    \includegraphics[width=\linewidth]{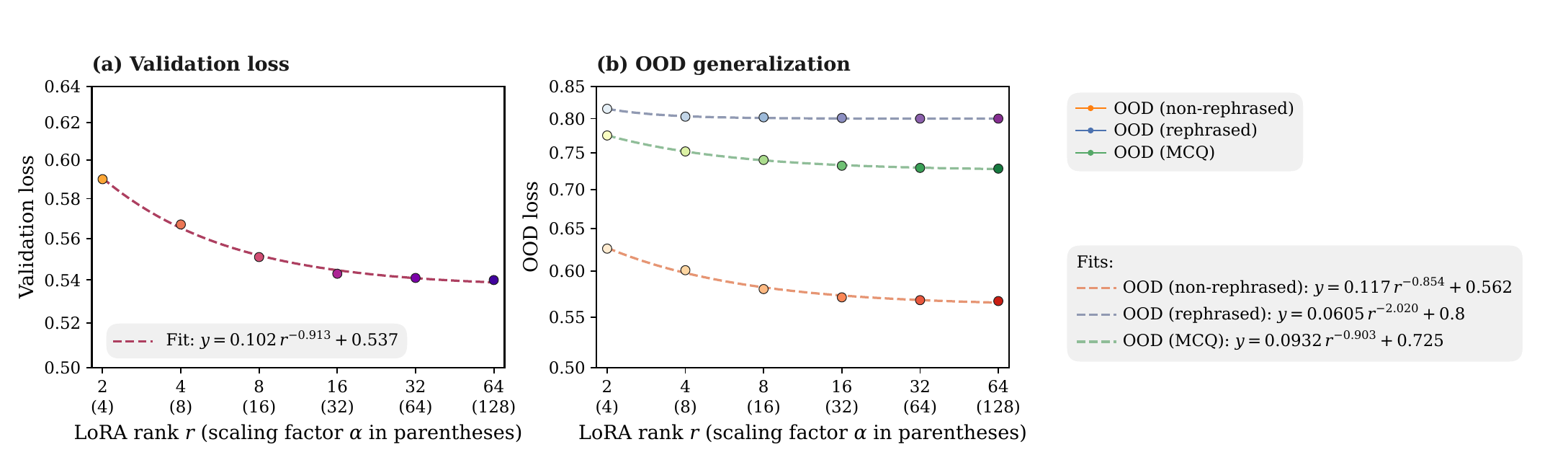}
    \caption{LoRA rank scaling results, showing final epoch loss vs.\ LoRA 
    rank $r$. Unlike other scaling axes, LoRA rank scaling exhibits a clear 
    saturating regime and is well-fit by a power-law with a constant offset: 
    $\mathcal{L}(r) = a \cdot r^b + c$. Fitted parameters: 
    validation $0.102 \cdot r^{-0.913} + 0.537$, OOD non-rephrased 
    $0.117 \cdot r^{-0.854} + 0.562$, OOD rephrased $0.0605 \cdot r^{-2.020} 
    + 0.800$, and OOD MCQ $0.0932 \cdot r^{-0.903} + 0.725$.}
    \label{fig:lora_rank_loglog}
\end{figure}

To complement the target-model-scale comparison in 
Section~\ref{sec:lora_target_scaling}, we also examine how LoRA finetuning 
scales with adapter capacity. We train LoRA finetuning baselines across 
ranks $r \in \{2, 4, 8, 16, 32, 64\}$ with scaling factor $\alpha = 2r$, 
holding all other hyperparameters fixed and using the same target model 
as the hypernetwork experiments. Figure~\ref{fig:lora_rank_loglog} shows 
the final epoch loss against $r$.

Unlike the other scaling axes, LoRA rank scaling exhibits a clear 
saturating regime and cannot be captured by a pure power-law. We therefore 
fit a power-law with an additive constant, $\mathcal{L}(r) = a \cdot r^b 
+ c$, where $c$ represents the asymptotic loss floor. The fitted floor 
$c$ is substantial in every metric (0.537 for validation, 0.562, 0.800, 
and 0.725 for OOD non-rephrased, rephrased, and MCQ respectively), meaning 
that increasing adapter capacity beyond a modest rank produces diminishing 
returns and cannot close the OOD gap to the hypernetwork. This saturating 
behavior contrasts with the target-model-scale results in 
Section~\ref{sec:lora_target_scaling}, where LoRA continues to improve 
without a visible plateau, indicating that target model capacity is the 
dominant scaling axis for LoRA-based knowledge injection.

\section{Loss Trajectories During Training}
\label{sec:training_trajectories}

In this section, we provide the full loss trajectories for all scaling 
experiments, plotting validation loss and OOD non-rephrased loss against 
cumulative training FLOPs. Using FLOPs as the x-axis rather than steps or 
epochs enables direct comparison across configurations that differ in 
per-example compute cost (see Appendix~\ref{sec:training_flops}). In all 
plots, darker colors correspond to larger configurations (i.e., wider, deeper, 
larger target model, or more facts). The monotonic ordering of the loss values by configuration 
is established early during training and largely maintained throughout, confirming that 
the scaling trends reported in Section~\ref{sec:experiments} are stable and 
not artifacts of training length.

\begin{figure}[h]
    \centering
    \includegraphics[width=\linewidth]{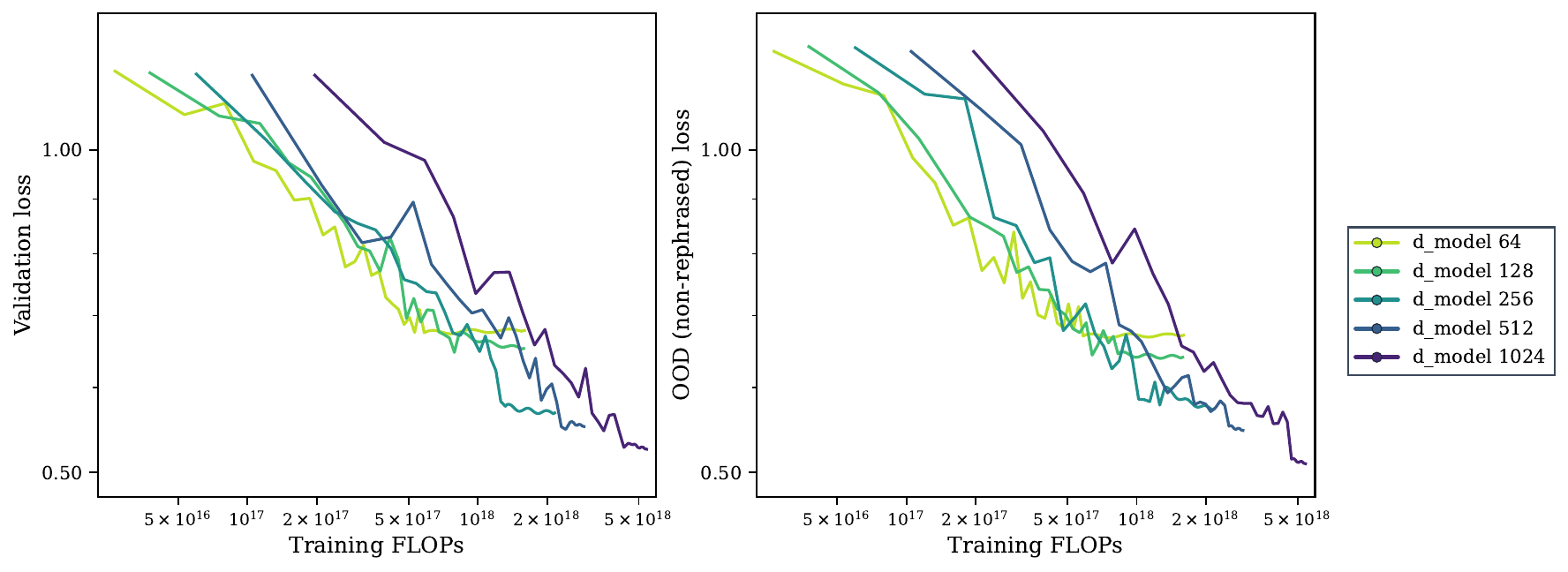}
    \caption{Loss trajectories for hypernetwork width scaling, showing 
    validation loss (left) and OOD non-rephrased loss (right) vs.\ cumulative 
    training FLOPs for $d_{\text{model}} \in \{64, 128, 256, 512, 1024\}$.}
    \label{fig:width_traj}
\end{figure}

\begin{figure}[h]
    \centering
    \includegraphics[width=\linewidth]{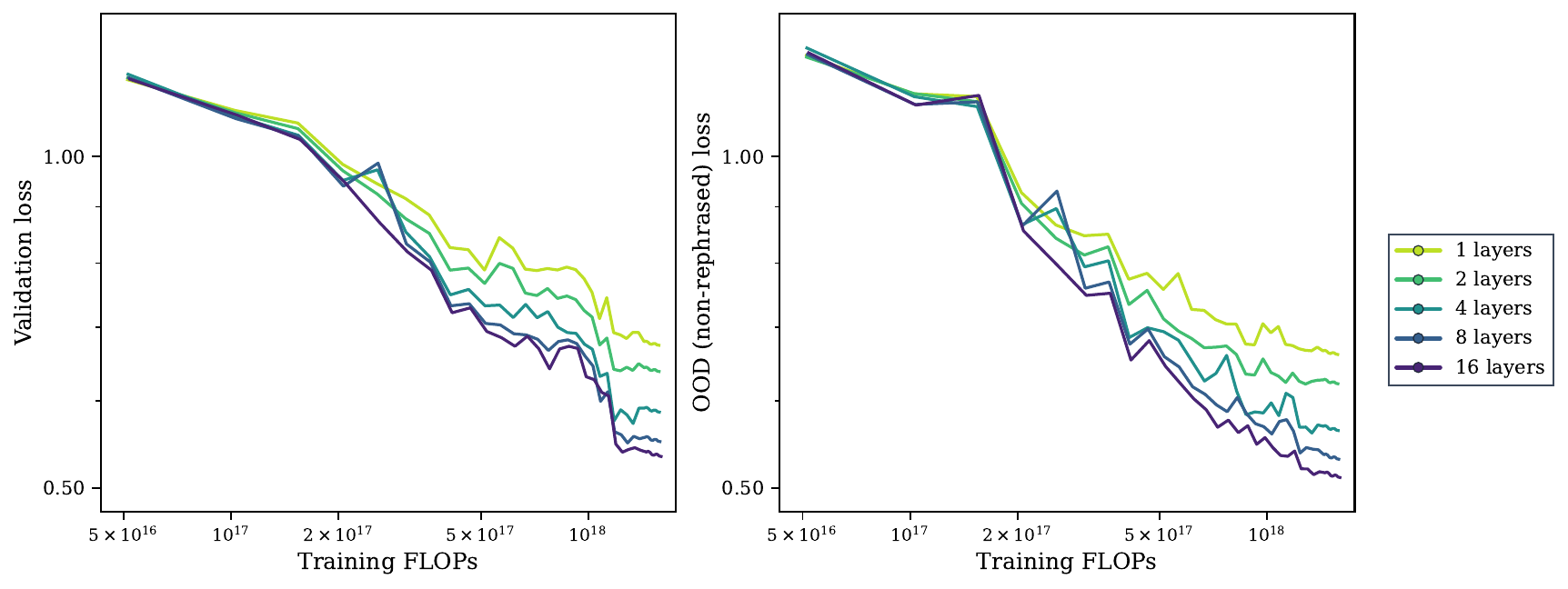}
    \caption{Loss trajectories for hypernetwork depth scaling, showing 
    validation loss (left) and OOD non-rephrased loss (right) vs.\ cumulative 
    training FLOPs for $L_{\text{HN}} \in \{1, 2, 4, 8, 16\}$. The narrow 
    FLOPs range reflects the near iso-compute nature of depth scaling 
    (see Table~\ref{tab:flops_depth}).}
    \label{fig:depth_traj}
\end{figure}

\begin{figure}[h]
    \centering
    \includegraphics[width=\linewidth]{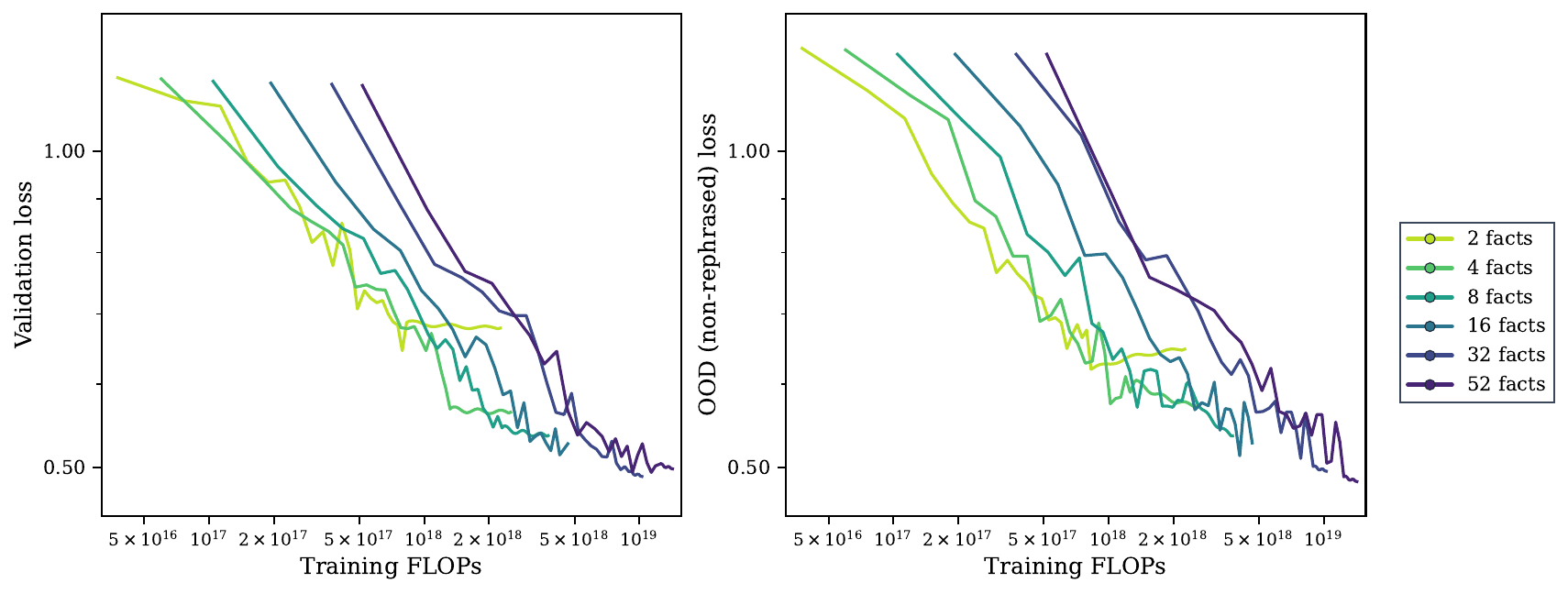}
    \caption{Loss trajectories for fact count scaling, showing validation 
    loss (left) and OOD non-rephrased loss (right) vs.\ cumulative training 
    FLOPs for $N_{\text{facts}} \in \{2, 4, 8, 16, 32, 52\}$. Configurations 
    with more facts reach higher total FLOPs due to the larger $n_h$ 
    (see Table~\ref{tab:flops_facts}).}
    \label{fig:facts_traj}
\end{figure}

\begin{figure}[h]
    \centering
    \includegraphics[width=\linewidth]{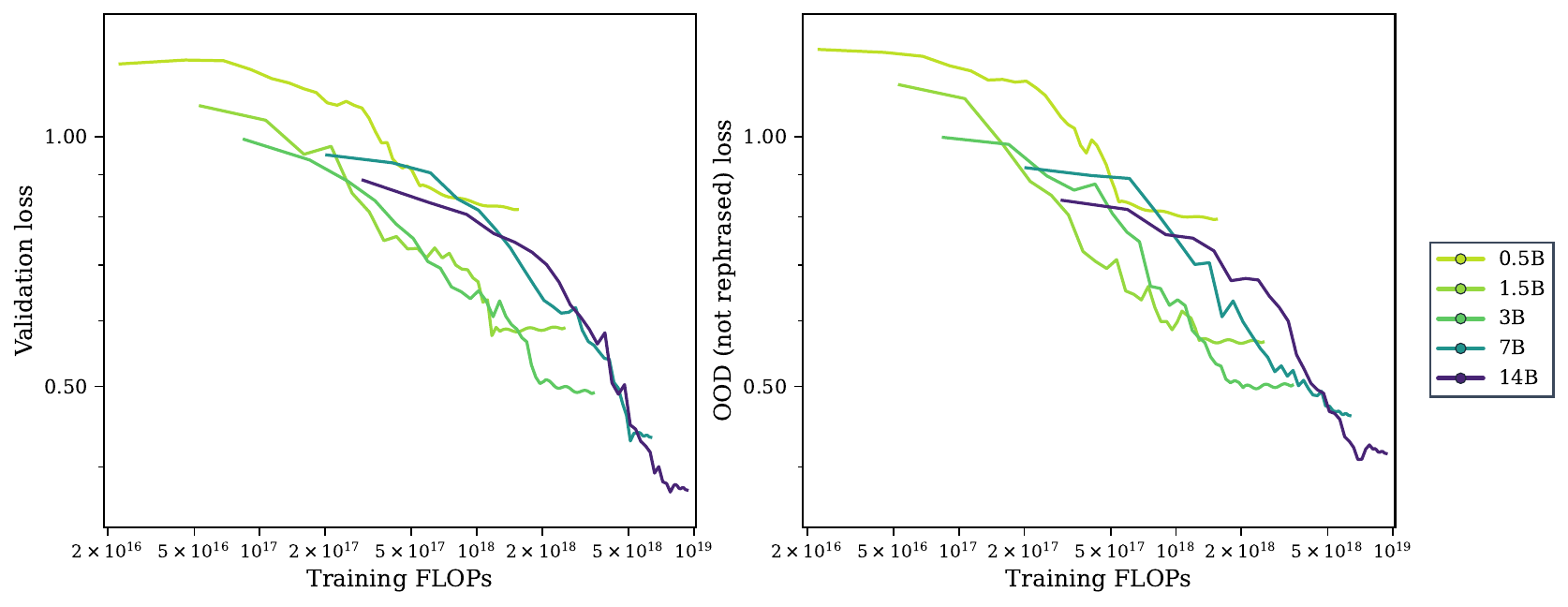}
    \caption{Loss trajectories for target model scaling, showing validation 
    loss (left) and OOD non-rephrased loss (right) vs.\ cumulative training 
    FLOPs for target model sizes 0.5B, 1.5B, 3B, 7B, and 14B. Larger target 
    models incur higher per-example FLOPs and therefore reach higher total 
    FLOPs at the same number of epochs.}
    \label{fig:target_traj}
\end{figure}
\end{document}